\documentclass[sigconf, nonacm]{acmart}

\newcommand\vldbdoi{10.14778/3718057.3718072}
\newcommand\vldbpages{1453 - 1465}
\newcommand\vldbvolume{18}
\newcommand\vldbissue{5}
\newcommand\vldbyear{2025}
\newcommand\vldbauthors{\authors}
\newcommand\vldbtitle{\shorttitle} 
\newcommand\vldbavailabilityurl{https://github.com/UIC-InDeXLab/Mining_U3Ms}
\newcommand\vldbpagestyle{empty} 
\usepackage{color}	
\usepackage{xspace}
\usepackage{graphicx}
\usepackage{color}
\usepackage{enumerate}
\usepackage{hyperref}
\usepackage[font={small,sf}]{caption}
\usepackage{floatflt}
\usepackage{lipsum}
\usepackage{multirow}
\usepackage{soul}
\sethlcolor{black!10}
\usepackage{mdframed}
\usepackage{amsmath,amsfonts}
\usepackage{comment}
\usepackage{subcaption}
\usepackage{enumitem}
\usepackage{csquotes}
\usepackage{tabularx}
\usepackage{tabularx}
\usepackage{enumitem}
\usepackage{tikz}
\usetikzlibrary{positioning}
\usepackage{mathtools}
\usepackage[a-2b]{pdfx}  
\usepackage[T1]{fontenc}
\usepackage{microtype}
\usepackage{bbm}
\usepackage{fnpct}  
\usepackage{tcolorbox}
\graphicspath{{./figures/}}
\usepackage{pgfplots}
\pgfplotsset{compat=1.18}  

\usepackage{algorithm}
\usepackage[noend]{algpseudocode}

\newcommand{\techrep}[1]{}
\newcommand{\submit}[1]{#1}

\DeclarePairedDelimiterX\set[1]\lbrace\rbrace{\,#1\,}

\usepackage{tikz}
\usetikzlibrary{positioning}

\newcommand*{\Reals}{\mathbb{R}}

\DeclareMathOperator*{\argmax}{arg\,max}

\newcommand{\stitle}[1]{\vspace{1mm}\noindent{\textbf{#1}}.}

\newcommand{\gee}{\mathbf{g}}
\newcommand{\dee}{\mathcal{D}}

\newcommand{\EX}{\mathbb{E}}

\newcommand{\at}[1]{{\tt \small #1}\xspace}

\newcommand{\dual}{\mathsf{d}}

\newcommand{\yooms}{Minoria\xspace}
\newcommand{\yoom}{Minoria\xspace}
\newcommand{\warmup}{Warm-up\xspace}
\newcommand{\raysweeping}{Ray-sweeping\xspace}

\newtheorem{theorem}{Theorem} 
\newtheorem{lemma}[theorem]{Lemma}
\newtheorem{property}{Property}
\newtheorem{definition}{Definition}
\newtheorem{problem}{Problem}
\newtheorem{example}{Example}

\DeclarePairedDelimiter\ceil{\lceil}{\rceil}
\DeclarePairedDelimiter\floor{\lfloor}{\rfloor}

\newtcolorbox{highlightbox}{
  colback=blue!10,
  colframe=blue!20,
  arc=0.5mm, 
  fonttitle=\bfseries,
  boxrule=0mm,
  boxsep=0mm,
  left=0mm,
  right=0mm,
  top=0mm,
  bottom=0mm
}

\newtcolorbox{examplebox}{
  colback=blue!10,
  colframe=blue!20,
  arc=2mm, 
  fonttitle=\bfseries,
  boxrule=0mm,
  boxsep=1mm,
  left=0mm,
  right=0mm,
  top=0mm,
  bottom=0mm
}

\newtcolorbox{defbox}{
  colback=orange!10,
  colframe=orange!20,
  arc=2mm, 
  fonttitle=\bfseries,
  boxrule=0mm,
  boxsep=1mm,
  left=0mm,
  right=0mm,
  top=0mm,
  bottom=0mm
}

\newtcolorbox{pbox}{
  colback=black!5,
  colframe=black!30,
  arc=2mm, 
  fonttitle=\bfseries,
  boxrule=0mm,
  boxsep=1mm,
  left=0mm,
  right=0mm,
  top=0mm,
  bottom=0mm
}
\usepackage{graphicx,epstopdf}
\begin{document}
\title{Mining the Minoria: Unknown, Under-represented, and Under-performing Minority Groups}
\titlenote{This work was supported in part by the National Science Foundation, Grant No. 2348919 and 2107290.}

\author{Mohsen Dehghankar}
\affiliation{%
  \institution{University of Illinois Chicago}
}
\email{mdehgh2@uic.edu}

\author{Abolfazl Asudeh}
\affiliation{%
  \institution{University of Illinois Chicago}
}
\email{asudeh@uic.edu}

\begin{abstract}
Due to a variety of reasons, such as privacy, data in the wild often misses the grouping information required for identifying minorities. On the other hand, it is known that machine learning models are only as good as the data they are trained on and, hence, may underperform for the under-represented minority groups. The missing grouping information presents a dilemma for responsible data scientists who find themselves in an unknown-unknown situation, where not only do they not have access to the grouping attributes but do not also know what groups to consider.

This paper is an attempt to address this dilemma. Specifically, we propose a minority mining problem, where we find vectors in the attribute space that reveal potential groups that are under-represented and under-performing. Technically speaking, we propose a geometric transformation of data into a dual space and use notions such as the arrangement of hyperplanes to design an efficient algorithm for the problem in lower dimensions. Generalizing our solution to the higher dimensions is cursed by dimensionality. Therefore, we propose a solution based on smart exploration of the search space for such cases. We conduct comprehensive experiments using real-world and synthetic datasets alongside the theoretical analysis. Our experiment results demonstrate the effectiveness of our proposed solutions in mining the unknown, under-represented, and under-performing minorities.
\end{abstract}

\maketitle

\pagestyle{\vldbpagestyle}
\begingroup\small\noindent\raggedright\textbf{PVLDB Reference Format:}\\
\vldbauthors. \vldbtitle. PVLDB, \vldbvolume(\vldbissue): \vldbpages, \vldbyear.\\
\href{https://doi.org/\vldbdoi}{doi:\vldbdoi}
\endgroup
\begingroup
\renewcommand\thefootnote{}\footnote{\noindent
This work is licensed under the Creative Commons BY-NC-ND 4.0 International License. Visit \url{https://creativecommons.org/licenses/by-nc-nd/4.0/} to view a copy of this license. For any use beyond those covered by this license, obtain permission by emailing \href{mailto:info@vldb.org}{info@vldb.org}. Copyright is held by the owner/author(s). Publication rights licensed to the VLDB Endowment. \\
\raggedright Proceedings of the VLDB Endowment, Vol. \vldbvolume, No. \vldbissue\ %
ISSN 2150-8097. \\
\href{https://doi.org/\vldbdoi}{doi:\vldbdoi} \\
}\addtocounter{footnote}{-1}\endgroup

\ifdefempty{\vldbavailabilityurl}{}{
\vspace{.3cm}
\begingroup\small\noindent\raggedright\textbf{PVLDB Artifact Availability:}\\
The source code, data, and/or other artifacts have been made available at \url{https://github.com/UIC-InDeXLab/Mining_U3Ms}.
\endgroup
}

\section{Introduction}
Specifying the scope of use of a dataset is an important step before sharing it~\cite{sun2019mithralabel}. Particularly, it is known that data-driven models are not reliable and may be unfair towards the groups that are not well represented in their (training) dataset~\cite{shahbazi2024reliability,shahbazi2024coverage}.
Therefore, it is important to identify the potential groups that are {\em under-represented} and for which the models {\em under-perform}.
However, {\em data in practice often misses to include demographic information of entities.} 
Even in cases where some demographic information are available, many others may be missing that is necessary for identifying representation bias~\cite{shahbazi2023representation}.
To further motivate this problem, let us consider the following example:

\begin{pbox}
    \begin{example}\label{ex:motivation}
        A data-sharing platform (e.g., \at{Chicago Open Data Portal}~\cite{ChiDataPort}) would like to responsibly specify the {\em under-represented} and {\em under-performing} groups in their shared data in order to limit their scope of use. However, their datasets either do not include grouping attributes (such as \at{race}) or only contain some of those.
        Moreover, targeting a comprehensive audit, they do not want to limit their scope to a small set of predefined groups. Instead, they want to be proactive in their detection process, even for the groups they might have missed to consider.
    \end{example}
\end{pbox}

For some cases such as image datasets, when a small and predetermined set of grouping attributes (e.g., \at{gender}) are missing, one may be able to use techniques such as crowdsourcing to identify under-represented groups based on those attributes~\cite{melika}. 
Our problem, however, is that the grouping attributes are not predetermined. 
In other words, we consider an {\em unknown unknown} situation where not only do we lack access to the grouping attributes but we also do not know the groups of interest.

To address this problem, we introduce the problem of mining the potential groups that are under-represented and under-performing, when the grouping attributes are unknown. We refer to this problem as {\em \yoom mining}.
To the best of our knowledge, this paper is {\em the first to introduce and study the \yoom mining problem}.
In simple terms, the main question we want to answer in this paper is, 
``\at{are there regions in data-space, representing  potential groups that are (a) under-represented, for which (b) the model under-performs}?'' 
For instance, aligned with Example~\ref{ex:motivation}, our experiments on a Crimes dataset (\S~\ref{sec:exp:chicagocrimes}) identified \at{North-side Chicago} as under-represented and under-performing. 
This information helps the model developers and data owners such as \at{Chicago Open Data Portal} to proactively identify a potential issue, decide if it is socially concerning,  and limit the dataset's scope-of-use accordingly.


Similar to any other data mining problem, multiple approaches may come to mind for addressing this problem.
For example, one can adapt a clustering technique (such as k-means clustering~\cite{macqueen1967some} or subspace and projective clustering ~\cite{moise2009subspace, parsons2004subspace,gullo2013projective}) to cluster the data into potential groups. 
However, as we shall show in our experiments (\S~\ref{sec:exp:2d:2}) such an approach fails to identify under-represented groups, simply because there are not enough instances from those groups in the dataset. 
This is a well-known drawback of the clustering methods, that their performance degrades substantially for detecting small clusters with insufficient data points~\cite{rodriguez2014clustering}.

Instead, in this paper, 
we propose a novel approach that involves identifying high-skew projections of data along a specific direction in the feature space.
Before outlining our technical contributions, we acknowledge that our approach is {\em only one of the possible solutions} for \yoom mining.
There may be other complementary and equally effective techniques that address this problem. We hope that this paper will pave the way for future research to explore alternative approaches.

\subsection{Technical Contributions}

We propose a \yoom mining approach  
that assists its users to mine unknown, under-represented, and under-performing groups
{\em as combinations of other features}.

Consider projecting data on a combination\footnote{We interchangeably use the term ``linear direction'' or simply ``direction'' for ``linear combination''.} of features, such that (i) the projected data is highly skewed (hence its tail indicating an under-represented group candidate) and (ii) the prediction loss at the tail is high.
Our mining task is to discover such cases and {\em return the under-performing samples} (e.g., the misclassified samples) in the tail to the human expert for further investigation to verify if those represent a socially-valid group. 


\begin{figure}
    \centering
    \includegraphics[width=0.5\linewidth]{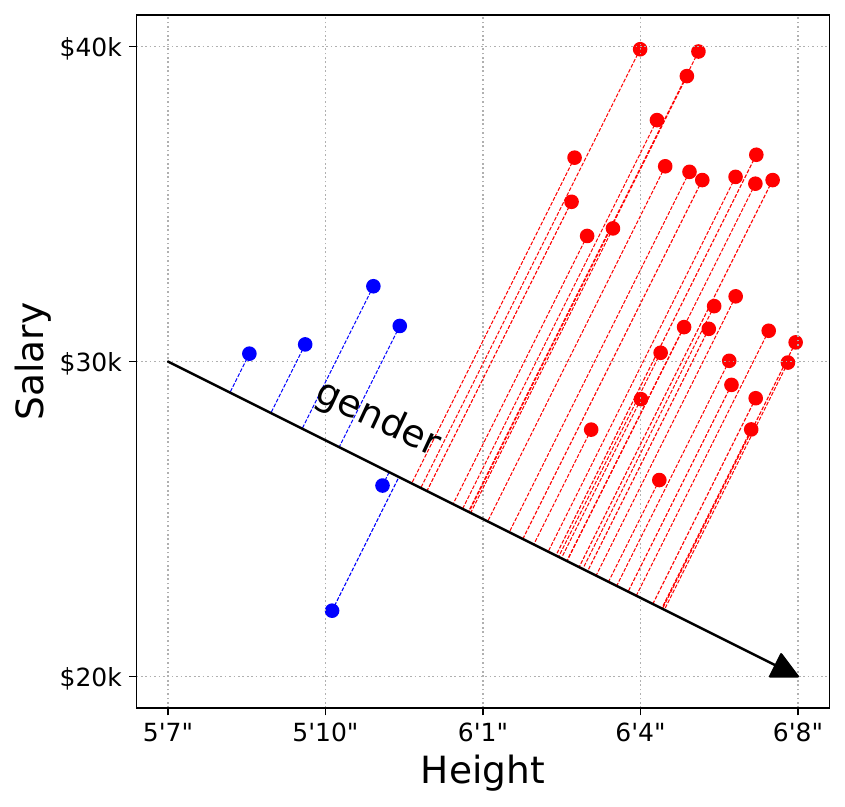}
    \vspace{-4mm}
    \caption{Toy example: \yoom mining using high-skew projections.}
    \label{fig:ex-fm}
    \vspace{-3mm}
\end{figure}

To better explain our approach, let us consider a toy example of a \at{basketball} dataset with several observation features $X$, including \at{height}
 and \at{salary}, and one target attribute ($y$) \at{performance-score}, where the goal is to predict $y$ using $X$.
 Let the \at{(height, salary)} values be as shown in Fig.~\ref{fig:ex-fm}. Our system discovers the black vector as a combination of \at{height} and \at{salary}, where the projection of values on it are highly skewed. 
 That is, after orthogonally projecting each point on the black vector, the distribution of the projected values is highly skewed. 
 In this example, the projections are specified with dashed line, while the point in the head and tail of the distribution colored in red and blue, 
 respectively.
 The points in the tail are considered as a {\em potential groups} that are under-represented.
 Next, we observe that the prediction performance is significantly less for this potential group. We then report the misclassified samples on the tail as a potential \yoom to the user for further investigation.
 In this example, the discovered group represents (mostly) the \at{female} athletes that are not well-represented in the dataset.


\stitle{Summary of contributions} 
In this paper, we introduce and formalize the \yoom mining problem.
We propose an approach for the problem based on the high-skew projections in the feature space that reflect performance-disparities at the projection tail.
It turns out that finding such projections is challenging as it requires identifying the median of each projection.
To address this challenge, we utilize computational geometry notions such as duality, $k$-th level of arrangement, and the arrangement skeleton, and define the concept of median regions~\cite{edelsbrunner1987algorithms}. We prove the sufficient conditions for finding the maximum skew within each median region, which enables developing an efficient \raysweeping algorithm for a constant number of dimensions. Our \raysweeping algorithm has {\em the same time complexity} as of the k-th level of arrangement enumeration.
This algorithm, however, suffers from the curse of dimensionality. Therefore, we propose practical solutions such as search-space discretization and smart exploration of the search space for large-scale high-dimensional settings.
Besides theoretical guarantees, our experiments on multiple datasets confirm the validity of our proposal and the performance of our algorithms.
For example, our experiments on a Chicago crime dataset discovered an under-preforming high-skew projection that (mostly) reflects the \at{white} group.

\section{Preliminaries}
\stitle{Notations} We use $\{.\}^n$ to show a set of size $n$, and $[n]$ for the set of integers $\{1,\cdots,n\}$.
We also show the inner product of two vectors $t$ and $f$ as $t^\top f$.

\stitle{Data Model}
We consider a dataset $\dee=\{t_i\}^n$ as a collection of $n$ sample points (aka tuples).  
Each sample $t_i\in \dee$ is defined as a tuple $t_i=\langle X,y \rangle$.
$X=\langle \mathbf{x}_1,\cdots,\mathbf{x}_d\rangle$,
is a $d$-dimensional vector of Real values that corresponds with a set of {\em observation} attributes, also known as features. The observation attributes are used for prediction.
For each sample $t_i\in\dee$, we use $\langle t_{i, 1}, \cdots, t_{i, d}\rangle$, to show its features.
The dataset also contains a {\em target attribute} (aka label) $y$.\footnote{The extension of our solutions for datasets with more than one target attribute is straightforward by solving the problem separately for each attribute $y_k$.}
The dataset $\dee$ is used to train a model $h_\theta(X)$ that predicts $y$ using the observation attributes.

\stitle{Group Membership} We consider a universe of (demographic) groups such as \at{male}, \at{female}, \at{Black}, \at{White}, etc., to which a subset of points belong to. For a group $\gee$, we use $\dee^\gee = \gee\cap \dee$ to specify the set of points that belong to it and $\dee^{!\gee} = \dee\backslash \dee^\gee$ as the points that do not belong to it.
A grouping (aka, sensitive) attribute separates the group $\gee$ from the others.
For example, the grouping attribute \at{gender} separates the \at{female} individuals ($\dee^{\mbox{f}}$) from others ($\dee^{!\mbox{f}}$).
We assume {\em the dataset does not include the grouping attributes}. 
Moreover, {\em the set of grouping attributes of interest is not predetermined}.
In other words, not only do we not know the grouping information, we do not also know what groups we should consider.

\subsection{Problem Definition}

\stitle{\yoom Mining}
Our objective is to mine the unknown, under-represented, and underperforming minority groups (referred to as \yooms). Specifically, given a dataset $\dee$, identify groupings of $\dee$ into $\dee^\gee$ and $\dee^{!\gee}$ such that the following conditions hold\footnote{Note that, slightly abusing the notation, we also use $\gee$ for the group candidates. A discovered group candidate may or may not express a socially valid group.}:
\begin{enumerate}[leftmargin=*]
    \item[(a)] 
    $\gee$ is under-represented in the dataset, i.e., $|\dee^\gee|\ll |\dee|$.
    \item[(b)]
    Predictions based on $\dee$ are not accurate for $\gee$. Formally, let $h_\theta$ be the optimal model that minimizes the expected loss $\EX[L_\dee(\theta)]$ on $\dee$.
    Then, for a given performance-disparity threshold $\tau$, $\EX[L_{\dee^\gee}(\theta)] - \EX[L_\dee(\theta)]\geq \tau$.
\end{enumerate}
We combine $X$ to form {\em proxy} attributes that identify \yooms.
Formally, we would like to find projections of points in $\dee$ from $\Reals^d$ to $\Reals$ such that the projection (almost) separates $\dee^\gee$ from $\dee^{!\gee}$.
Furthermore, to satisfy condition (a), we are interested in finding projections with {\em high skew} on $\dee$, indicating an under-represented group candidate at the tail of the projection. 

Having identified such a group candidate $\gee$, we return the under-performing samples in $\gee$ (the tail of a high-skew projection) to the human expert for further investigation.

Without loss of generality, we focus on linear projections in this paper and use \emph{Pearson's median skewness}~\cite{doane2011measuring} for computing the skew on a projection.
The {Pearson's median skewness} of a set of values $V = \{v_i\}^n$, where $v_i \in \Reals$ is computed as: 

\vspace{-4mm}
\begin{align}
    skew(V) = \frac{3(\mu(V) - \nu(V))}{\sigma(V)}
\end{align}

Where $\mu(V)$, $\nu(V)$, and $\sigma(V)$ are mean, median, and standard deviation of $V$. For a unit vector $f$, we define $\dee_f$ as the projection of all points on $f$: 

\vspace{-8mm}
\begin{align}
   \hspace{20mm} \dee_f = \{t_i^\top f ~|~ t_i \in \dee\}
\end{align}

Then, the {\it unit} vector $f^*$ with the highest skew is defined as $f^* = \argmax_{f^\top f = 1, f \in \Reals^d} skew(\dee_f)$.



Since we would like to identify the \yoom candidates for further investigation, instead of returning the global optimum, i.e., the max-skew projection, we are interested in finding a top-$\ell$  high-skew optima. Putting everything together, we define the \yoom mining problem as follows:

\begin{pbox}
\begin{problem}[\yoom Mining]\label{problem}
    Given a dataset $\dee$ of $n$ samples, each containing a
    $d$-dimensional observation vector $X$ and a target attribute $y$, and the performance-disparity threshold $\tau$, 
    find the unit vectors $\{f_1,\cdots,f_\ell\}$ as

    \vspace{-4mm}
    \begin{align*}
       \hspace{10mm} \{f_1,\cdots,f_\ell\} &= \underset{f^\top f = 1, f \in \Reals^d}{\mbox{top-}\ell} skew(\dee_f)
    \end{align*}
    such that for all $ i\in[\ell],~ \EX[L_{\dee^{\gee_i}}(\theta)] - \EX[L_\dee(\theta)]\geq \tau$.\\
    {\small top-$\ell$ refers to the top $\ell$ local optima (for a small constant $\ell$) with the maximum skew, $\theta$ is the parameter of the model 
    $h_\theta$ that minimizes the expected loss on $\dee$, and $\gee_i$ is the $p$-percentile tail of $\dee_{f_i}$.} 
    \hfill$\square$
\end{problem}
\end{pbox}

Addressing Problem~\ref{problem} is challenging as it requires identifying the median of each possible projection while there are infinite possible projections defined using different vectors $f$.
To tackle this challenge, we provide a geometric dual transformation of $\dee$ in the next section that helps us design our efficient algorithm in \S~\ref{sec:2d}.

\section{Geometric Interpretation}\label{sec:geo:1}
In this section, we present the geometric interpretation of data that forms the foundation of our solutions in later sections.
First, in Section~\ref{sec:geo:1}, we provide a notion of duality~\cite{edelsbrunner1987algorithms}, where samples are presented as hyperplanes in a geometric space and prove some valuable characteristics of this transformation. 
We then consider the arrangement~\cite{edelsbrunner1987algorithms} of dual hyperplanes to identify high-skew projections efficiently. 

\subsection{Dual-Space Transformation}\label{sec:geo:1}

It is popular to present data in {\em primal space}, where each sample is shown as a point in $\Reals^d$.
Instead, we use the following {\em dual space} transformation ~\cite{chester2014computing,asudeh2019rrr} that the data points as a set of hyperplanes. 

\stitle{The dual space}  
Let $t_i = \langle t_{i_1}, t_{i_2}, ..., t_{i_d} \rangle$ be a tuple in the dataset $\dee$.
We define the dual representation of $t_i$, as the hyperplane $\dual(t_i)$ in $\Reals^d$ ($\{x_1, \cdots, x_d\}$ are the $d$ axes of the dual space):

\vspace{-5mm}
\begin{align}
    \dual(t_i): t_{i_1} x_1 + t_{i_2} x_2 + ... + t_{i_d} x_d = 1 \label{Duality}
\end{align}\vspace{-5mm}

\begin{figure*}[t]
\centering
\begin{subfigure}[t]{.32\textwidth}
  \centering
  \includegraphics[width=.9\linewidth]{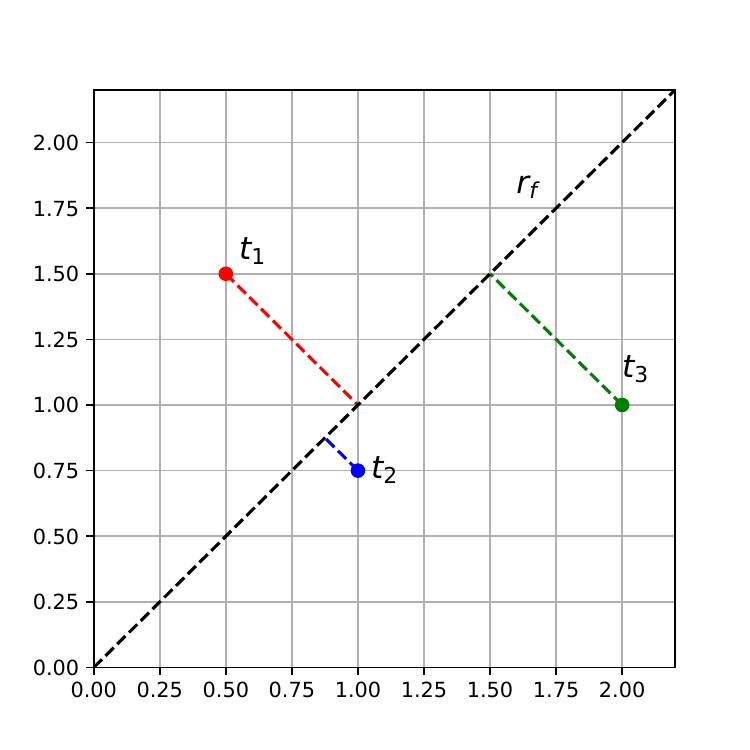} 
  \vspace{-5mm}\caption{Primal space: $[t_2, t_1, t_3]$ shown as points, along with their projection on the ray $r_f$.}
  \label{fig:toy1:primal}
\end{subfigure}\hfill
\begin{subfigure}[t]{.32\textwidth}
  \centering
  \includegraphics[width=.9\linewidth]{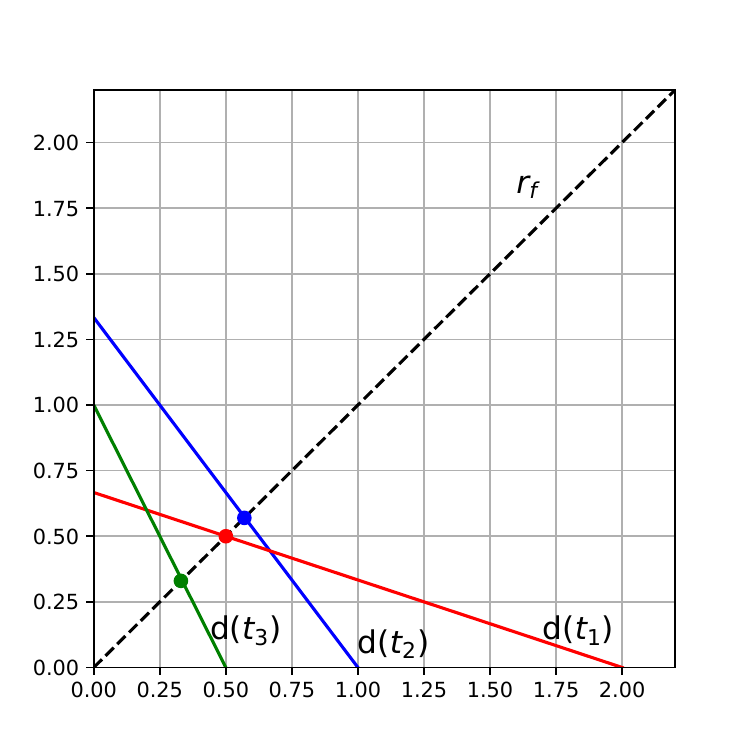}
  \vspace{-5mm}\caption{Dual space: The intersection of dual hyperplanes of the points with a ray $r_f$. 
  }
  \label{fig:toy1:dual}
\end{subfigure}\hfill
\begin{subfigure}[t]{.32\textwidth}
  \centering
  \includegraphics[width=.9\linewidth]{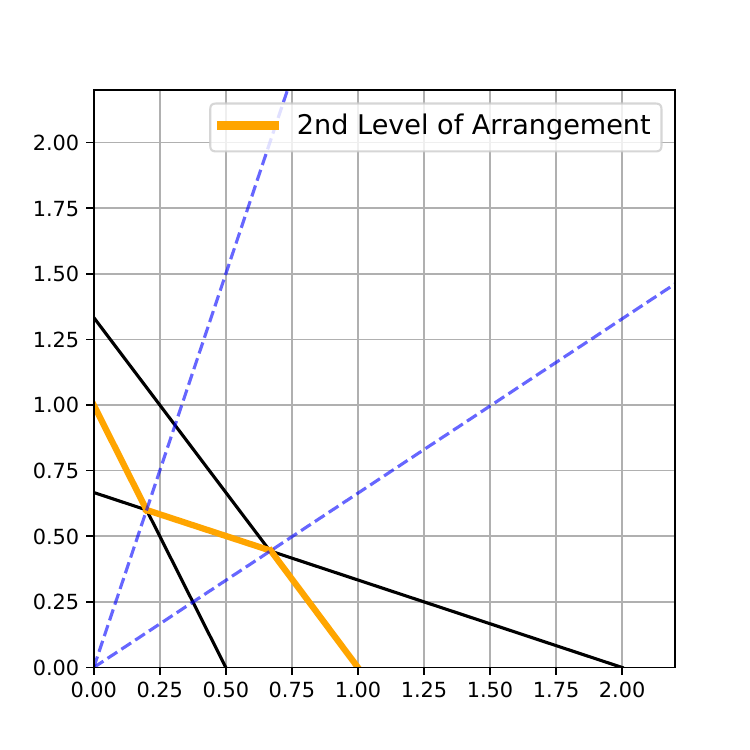}
  \vspace{-5mm}\caption{2nd level of arrangements in the dual space highlighted as the orange line segments.
  }
  \label{fig:toy1:level}
\end{subfigure}
  \vspace{-3mm}
\caption{The illustration of the toy dataset $\dee=\{t_1\langle .5,1.5 \rangle, t_2\langle 1,.75 \rangle, t_3\langle 2,1 \rangle\}$ in the primal space, the dual space, and the 2nd level of the arrangement in the first quadrant.
The order of projection in Fig.~\ref{fig:toy1:primal} is the reverse of the intersection of dual hyperplanes with $r_f$ ($[\dual(t_3), \dual(t_1), \dual(t_2)]$) in Fig.~\ref{fig:toy1:dual}.
In Fig.~\ref{fig:toy1:level}, the dotted blue lines indicate the boundaries of the median regions (the change in the line segment indicates a change in the median (2nd point) of $\dee_f$).
}\label{fig:toy1}
\vspace{-4mm}
\end{figure*}

As a running example in this section, consider a toy dataset $\dee=\{t_1\langle .5,1.5 \rangle, t_2\langle 1,.75 \rangle, t_3\langle 2,1 \rangle\}$. Fig.~\ref{fig:toy1:primal} shows $\dee$ as points in the primal space. In Fig.~\ref{fig:toy1:dual}, every $t_i$ is represented as a line in the dual space. For instance, the point $t_1\langle .5,1.5 \rangle$ is shown as the line $\dual(t_1): 0.5 x_1+1.5 x_2 = 1$.

Similarly, let $H$ be a hyperplane in the primal space, specified with a set of $d$ points $p_1,\cdots,p_d$.
The dual presentation of $H$ is the point $\dual(H)$, specified as the intersection of $\dual(p_1),\cdots, \dual(p_d)$.
In the following, we review some properties of this dual space transformation and its relationship with the primal space.
For ease of explanation, for every unit vector $f$, we define $\vec{r_f}$ as the origin-anchored ray that passes through the point $f$.

\begin{property}\label{lem:primal_dual} 
    Consider a point $t \in \Reals^d$ and a hyperplane $H$ in the primal space. $t$ lies on the hyperplane $H$ if and only if the point $\dual(H)$ lies on the hyperplane $\dual(t)$. 
\end{property}

Let $\vec{r_f}$ be the origin-anchored ray passing through $\dual(H)$ in the dual space.
Also, let $h$ be the intersection of $H$ with the origin-anchored vector that is orthogonal to it in the primal space.
Then, $H$ is defined as the set of points $\{x : x^{\top} h = |h|\}$. As a result, $\dual(H) = h / |h|$. This holds because $H = \{x | \dual(H)^{\top} x = 1\}$. Then, the unit vector $f = \frac{h}{|h|}$, defining the ray ${\vec r_f}$, is the unit vector of $H$.

\begin{property}\label{lem:last_dual_property}
Let $t$ be a point and $f$ be a unit vector in $\Reals^d$. 
The intersection of $\vec{r_f}$ and $\dual(t)$ is $f/(t^\top f)$. 
\end{property}

Property~\ref{lem:last_dual_property} holds because the
    intersection of $\dual(t)$ and $\vec{r_f}$ can be shown as $\alpha f$ for a scalar value $\alpha \in \Reals$. Let $\alpha f$ intersect with the hyperplane $\dual(t)$, so $\alpha f$ satisfies Equation \ref{Duality}. Then, 

\vspace{-3mm}
    \begin{align}
        t^\top (\alpha f) = 1 \Leftrightarrow \alpha t^\top f = 1 \Leftrightarrow \alpha = \frac{1}{t^\top f} 
    \end{align}

As a result of Property \ref{lem:last_dual_property}, for a pair of points $t_i$ and $t_j$, $t_i^\top f \leq t_j^\top f$ if and only if the intersection of $\dual(t_i)$ lies further than $\dual(t_j)$ on the ray $\vec{r_f}$. This observation will result in the following Lemma.

\begin{lemma}\label{DualTheorem}
    The ordering of points in $\dee_f$ is the same as the reverse order of intersections of $\dual(t_i)$ hyperplanes with $\vec{r_f}$. Hence, the dual-space transformation preserves the order of projected points.
\end{lemma}

\vspace{-1mm}
An illustration of these observations is shown in Fig.~\ref{fig:toy1:primal} and Fig.~\ref{fig:toy1:dual} for the direction $f=\langle \frac{\sqrt{2}}{2}, \frac{\sqrt{2}}{2} \rangle$.
As shown in Fig.~\ref{fig:toy1:primal}, the projected points on $f$ are ordered as $\{t_2, t_1, t_3\}$.
One can confirm that the reverse ordering of the dual-line intersections with $\vec{r_f}$ in Fig.~\ref{fig:toy1:dual} is also the same.

\vspace{-2mm}
\subsection{Median Region}

A key concept that helps us to solve Problem~\ref{problem} efficiently is the 
computational geometry notion of {\em $k$-th level of arrangement}~\cite{edelsbrunner1987algorithms}.
Consider the dual hyperplanes $\mathcal{H} = \{\dual(t_1),\cdots,\dual(t_n)\}$. The dissection of the dual space with the hyperplanes is called the {\em arrangement} $\mathcal{A}$ of $\mathcal{H}$, which we simply call the arrangement.
The intersection of $\delta\in[d]$ hyperplanes in $\mathcal{H}$ forms a ($d-\delta$)-face of the arrangement. In particular, the intersection of $d$ hyperplanes is a (0)-face (a point), referred to as a {\em vertex} of the arrangement, while the (1)-faces (line segments) connecting two vertices are called the {\em edges}. We call two vertices that share an edge as {\em neighbors}.
Please refer to~\cite{edelsbrunner1987algorithms} for more details about the arrangement.

Now consider the origin-anchored ray $\vec{r_f}$ in the dual space and the intersections of the dual hyperplanes with it.
Based on Lemma~\ref{DualTheorem}, the $k$-th closest intersection to the origin specifies the tuple with the $k$-th largest value in $\dee_f$. We define the universe of all such intersection points for all vectors $f$ as the $k$-th level of arrangement, referred as $\mathcal{A}_k$.
In $\Reals^2$, we show $\mathcal{A}_k$ as the ordered list of the vertices (the intersection points) of the $k$-th level, in counter-clock-wise (from $x$-axis to $y$) or clock-wise.

For example, Fig.~\ref{fig:toy1:level} shows (in orange) the
2nd level of an arrangement of 3 (toy) tuples in the dual space in $\mathbb{R}^2$.
In this example, $\mathcal{A}_2=\langle 0, \times_{1,2},\times_{1,3}, \frac{\pi}{2}\rangle$, where $\times_{i,j}$ is the intersection between $\dual(t_i)$ and $\dual(t_j)$.
One can see that the $k$-th level is a chain of line segments forming a non-convex shape. In addition, the set of top-$k$ tuples in each line segment is fixed and is different from the neighboring line segments. Moreover, {\em when $k=\frac{n}{2}$, the $k$-th level of the arrangement tracks the median for any possible direction $f$}.
Specifically, the intersection of $\mathcal{A}_{\frac{n}{2}}$ with a ray $\vec{r_f}$ is a line segment, specifying the median of $\dee_f$.
Evidently, the median of all rays intersecting with the same line segment in $\mathcal{A}_{\frac{n}{2}}$ is the same. Following this observation, we define the notion of median regions as follows.

\vspace{-2mm}
\begin{pbox}
\begin{definition}[Median Region]\label{def:mregion}
    A median region is the range of rays $\vec{r_f}$, for which the median of the projection $\dee_f$ is the same.
\end{definition}
\end{pbox}
\vspace{-1mm}


The partitioning of the origin-anchored rays in the dual space according to the $(d-1)$-dimensional simplexes (line-segments in 2D) of $\mathcal{A}_{\frac{n}{2}}$ provides the set of all median regions. Specifically, in 2D each median region is a range that begins/ends at the beginning/end of each line segment (e.g., see Fig.~\ref{fig:toy1:level}).
Moreover, since within every median region, no dual hyperplane intersects with the median simplex, the set of tuples with projection values being larger (resp. smaller) than the median does not change.
For example, in Fig.~\ref{fig:toy1:level}, consider the median region between the x-axis and the first (bottom) blue ray. 
$\dual(t_2)$ is the median line-segment. 
The intersection of $\dual(t_3)$ with any ray in this region remains below the median (closer to the origin). Similarly, the intersection of $\dual(t_1)$ above the median.

Using these observations, one can write a linear program for every median region to find the maximum skew within a median region. 
Instead, we introduce Theorem~\ref{median_region_theorem} as a key observation that enables a more efficient algorithm in Section~\ref{sec:2d}.
Let $\mu(\dee)$ be the vector in $\Reals^d$ that contains the mean of $\dee$, i.e., $\mu(\dee)=\frac{1}{n}\sum_{i=1}^n t_i$.
We define \hl{$q_i = t_i - \mu(\dee)$}.
In other words, $q_i$'s are the mean-normalized tuples of $\dee$. Let $Q$ be a $d \times n$ matrix where each column of $Q$ is $q_i^\top$. 
For a unit vector $f$, let $t_{m_f}$ be the median of $\dee$ after projection on $f$.
In other words, $t_{m_f}^\top f$ is the median of $\dee_f$.
Theorem~\ref{median_region_theorem} helps us to find the high skew direction vector $f^*$ efficiently\submit{\footnote{Due to the space limitations, the proof is provided in the technical report~\cite{technicalreport}.}}:

\begin{pbox}
    \begin{theorem}\label{median_region_theorem}
    For a given median region with median $t_{m_f}$, the max skew direction $f^*$ either falls in the boundaries of this median region or it is equal to $\frac{(Q Q^\top)^{-1} q_{m_f}}{|(Q Q^\top)^{-1} q_{m_f}|}$.
\end{theorem}
\end{pbox}

\vspace{-3mm}
\section{\yoom Mining in 2D}\label{sec:2d}
\vspace{-1mm}

In this section, we start by studying the problem in $\Reals^2$ -- referred to as the 2D case.
Before providing the technical details, we would like to clarify that in the 2D case, $\dee$ can have more than two observation attributes (used for prediction). However, the projection directions are considered with respect to a pair of observation attributes called attributes of interest.
For example, in our experiments in \S~\ref{sec:exp}, we use \at{longitude} and \at{latitude} as the attributes of interest for \yoom mining in a crime dataset.

Recall that our goal is to find a set of high-skew directions. 
Let $\mathcal{H} = \{\dual(t_i) | t_i \in \dee\}$ be the set of dual hyper-planes in the dual space and let $\mathcal{E}$ be an oracle that returns $\mathcal{A}_{\frac{n}{2}}$ (the ordered list of vertices in the $\frac{n}{2}$-th level of arrangement of $\mathcal{H}$) in $T_{enum}$ time. 

Given $\mathcal{A}_{\frac{n}{2}}$, we can iterate through the neighboring median regions by iterating through the elements (vertices) of this ordered list.
Based on Theorem ~\ref{median_region_theorem}, within each median region, the high-skew direction is either at the boundary points (two vertices of the line segment) or at a specific direction, which can be calculated in $O(d)$. 
Since this time is dominated by the time taken to calculate the skew value of the directions corresponding to the boundary points, we can ignore it while analyzing the complexity of our algorithm.

In the second phase of our algorithm, we should iterate through the neighboring median regions and calculate the skew value of the directions of the boundaries of these regions. We want to return the top-$\ell$ high-skew directions as the \yoom candidates. In order to do that, we use a Max-Heap data structure to push these directions based on their skew value. We define $T_{skew}$ as the running time of this phase. 

In the end, we pop the top-$\ell$ directions from the heap, evaluate the model on the tail of these directions, and check the error on the tail to see if it represents an under-represented group candidate. 
A visualization of this approach is provided in Figure~\ref{fig:ray-sweep-toy}. 

Let $T_{pop}$ be the time complexity of the final step.
Then, the overall time complexity of the algorithm is $T = T_{enum} + T_{skew} + T_{pop}$.

Having discussed the overall description of the 2D solution, let us next explain the oracle $\mathcal{E}$ for enumerating $\mathcal{A}_{\frac{n}{2}}$. 

\vspace{-3mm}
\subsection{Enumerating the $k$-th Level} 
\vspace{-1mm}

Our efficient algorithm (\S~\ref{sec:2d:ray}) takes $\mathcal{A}_{\frac{n}{2}}$, the $\frac{n}{2}$-th level in an ordered list, as the input. 
The $k$-th level of arrangement in $\Reals^2$ has a complexity (the number of its vertices) of $m=O(n.k^\frac{1}{3})$~\cite{dey1997improved}.
There are efficient deterministic and randomized algorithms for enumerating the $k$-th level.
The SOTA deterministic algorithm is by Timothy Chan~\cite{chan1995output}. It enumerates $\mathcal{A}_{k}$ as an ordered list in $O(n \log m + m \log^{1 + \alpha} n)$, where $\alpha$ is a small positive constant, $n$ is the number of hyperplanes, and $m$ is the number of vertices in the $k$-th level of arrangement.
The randomized incremental algorithm performs this traversal in $O(m + n \log n)$~\cite{chan1999remarks}.

\vspace{-3mm}
\subsection{The \warmup Algorithm}\label{sec:2d:warmup}
\vspace{-1mm}
So far, we outlined the key steps in our algorithmic framework. 
Following these steps, the \warmup algorithm\submit{\footnote{Please find the pseudo-code of the \warmup algorithm in the technical report~\cite{technicalreport}.}} first enumerates $\mathcal{A}_{\frac{n}{2}}$. It then 
computes the median by evaluating all projections along a specified direction. Finally, it identifies and retrieves the top skewed elements. \techrep{Algorithm~\ref{alg:2d:naive} illustrates this process.}

\vspace{-2mm}
\paragraph{Complexity Analysis}
The \warmup algorithm \techrep{(Algorithm~\ref{alg:2d:naive})} conducts the three aforementioned steps separately.
In the first step, it transforms the dataset into the dual hyperplanes and calls the oracle $\mathcal{E}$ to enumerate $\mathcal{A}_{\frac{n}{2}}$.
As a result, the first step takes $T_{enum} = O(n \log (n k^{\frac{1}{3}}) + n k^{\frac{1}{3}} \log^{1 + \alpha} n) = O(n^\frac{4}{3} \log^{1 + \alpha} n)$ for the deterministic algorithm, or $T_{enum} = O(n k^{\frac{1}{3}} + n \log n) = O(n^\frac{4}{3})$ when using the randomized incremental algorithm (note that $k=\frac{n}{2}$ here).
In the second step, for each vertex in $\mathcal{A}_\frac{n}{2}$, it calculates the $skew$ value and adds it to a max-heap. Calculating the skew requires calculating the mean, median, and standard deviation of points projected on the direction $f$ and takes $O(n)$. 
Inserting into a max-heap using a Fibonacci heap takes a constant time. As a result, $T_{skew} = O(n^2 k^{\frac{1}{3}}) = O(n^{\frac{7}{3}})$.
In the third step, the \warmup algorithm pops the high skew directions from the heap and calculates the model's loss on the tail of each direction. Popping from a max-heap takes an amortized time of $O(\log n)$. Calculating the model accuracy would take $O(n)$.
Hence, assuming that the number of directions evaluated before returning the top-$\ell$ is a small constant, $T_{pop} = O(n \,\log n)$.

As a result, the time complexity of $2D$\warmup is dominated by $T_{skew}$ and is equal to 
$T = T_{enum} + T_{skew} + T_{pop} = O(n^{\frac{7}{3}})$. 

\techrep{
\begin{algorithm}[!tb]
\caption{\warmup Algorithm for 2D}\label{alg:2d:naive}
\begin{algorithmic}[1]
\Procedure{2D\warmup}{$\dee, \theta$}
\State $\mathcal{H} \gets Dual(\dee)$
\State $\mathcal{A}_{\frac{n}{2}} \gets \mathcal{E}(\mathcal{H},\frac{n}{2})$\Comment{Ordered list of vertices in $\frac{n}{2}$th level of arrangement.}
\State $Heap \gets [~]$\Comment{Initialize a max heap.}
\For{$v$ in $\mathcal{A}_{\frac{n}{2}}$}
\State $f \gets v / ||v||$\Comment{Get the direction of point $v$.}
\State $skew \gets GetSkewNaive(\dee, f)$
\State $Heap.push(\langle f, skew\rangle )$\Comment{Push $f$ to a max heap based on the skew value.}
\EndFor
\State $Output \gets [~]$
\While{$Output$ size is less that $l$}
\State $f \gets Heap.pop()$\Comment{Get a candidate direction.}
\State $T \gets \textit{p-}tail(f)$\Comment{$T$ is a candidate {\it \yoom} subset of $\dee$.}
\If{$L_{T}(\theta) - L_{\dee}(\theta) > \tau$}
$Output.append(f)$
\EndIf
\EndWhile
\State \textbf{return} $Output$\Comment{candidate {\it \yoom} directions.}

\EndProcedure
\Procedure{GetSkewNaive}{$\dee, f$}
\State \textbf{return} $3(\mu(\dee_f) - \nu(\dee_f)) / \sigma(\dee_f)$\label{skew_naive_line}
\EndProcedure
\end{algorithmic}
\end{algorithm}
}

\subsection{The \raysweeping Algorithm}\label{sec:2d:ray}

The \warmup algorithm is super-quadratic and not efficient.
Therefore, in this section, we propose the \raysweeping algorithm that has {\em the same time complexity} as of the enumeration of $\mathcal{A}_{\frac{n}{2}}$.
Similar to the \warmup algorithm, the \raysweeping algorithm also follows the three steps of (1) enumerating $\mathcal{A}_{\frac{n}{2}}$, followed by (2) computing the medians and (3) identifying the high-skew projections. 
However, unlike \warmup, it leverages our earlier observation to compute medians more efficiently by sweeping a ray counterclockwise from the x-axis to the y-axis.
At a high level, it maintains some aggregates while sweeping over $\mathcal{A}_{\frac{n}{2}}$, which enables updating the skew values in {constant time}.

\begin{figure*}[t] 
    \centering
    \begin{subfigure}[t]{0.24\textwidth} 
        \centering
        \includegraphics[width=\textwidth]{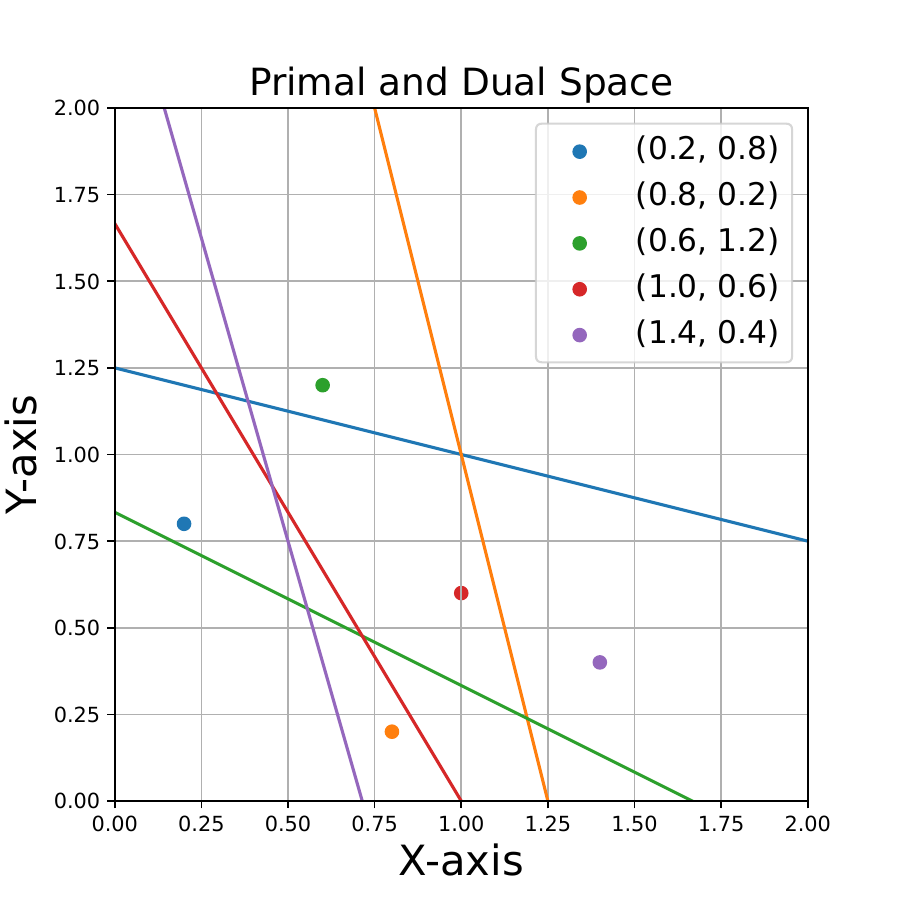}
        \vspace{-5mm}\caption{An example of $5$ points and their dual lines in the same plane.}
        \label{fig:figure1}
    \end{subfigure}
    \hfill
    \begin{subfigure}[t]{0.24\textwidth} 
        \centering
        \includegraphics[width=\textwidth]{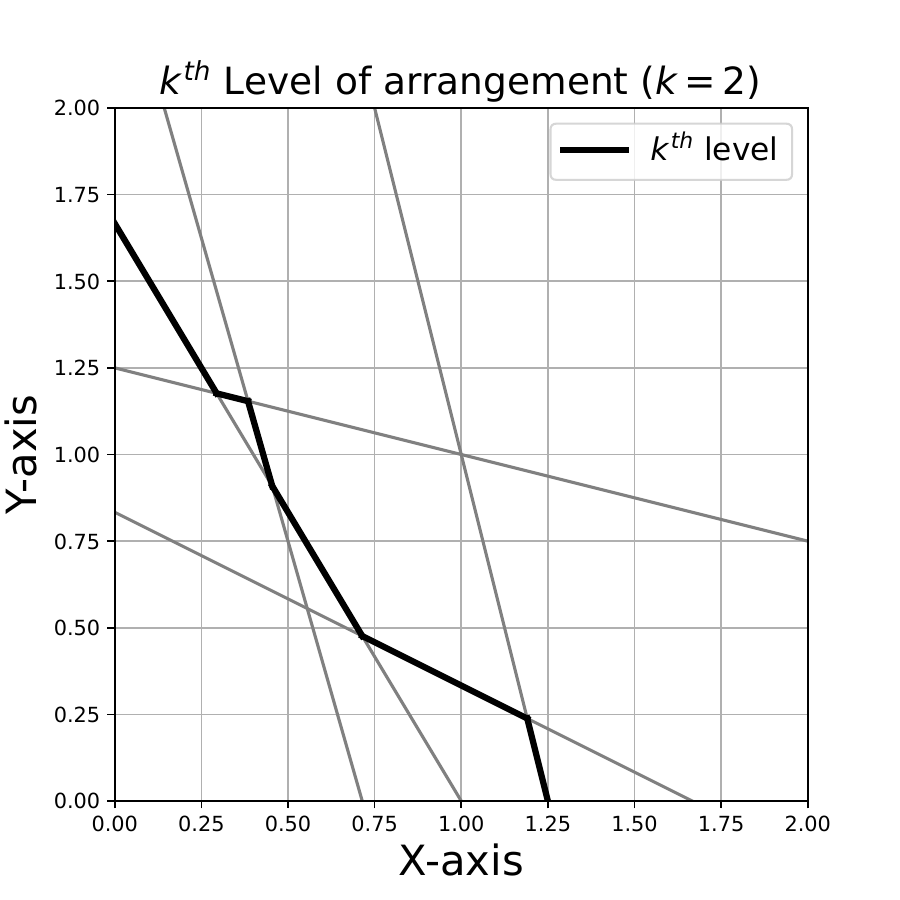}
        \vspace{-5mm}\caption{The $k$-th level of arrangement for $k = \frac{n}{2}$, representing the median points.}
        \label{fig:figure2}
    \end{subfigure}
    \hfill
    \begin{subfigure}[t]{0.24\textwidth} 
        \centering
        \includegraphics[width=\textwidth]{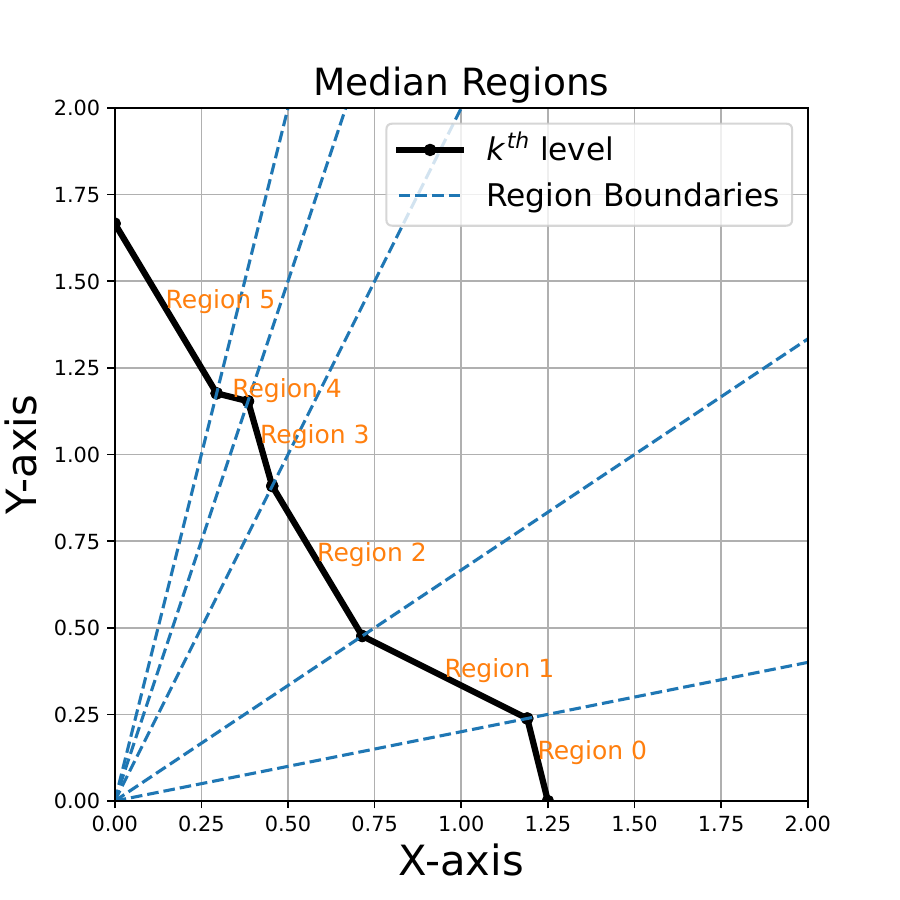}
        \vspace{-5mm}\caption{The median regions and their boundaries for a ray starting from the origin.}
        \label{fig:figure2}
    \end{subfigure}
    \hfill
    \begin{subfigure}[t]{0.24\textwidth} 
        \centering
        \includegraphics[width=\textwidth]{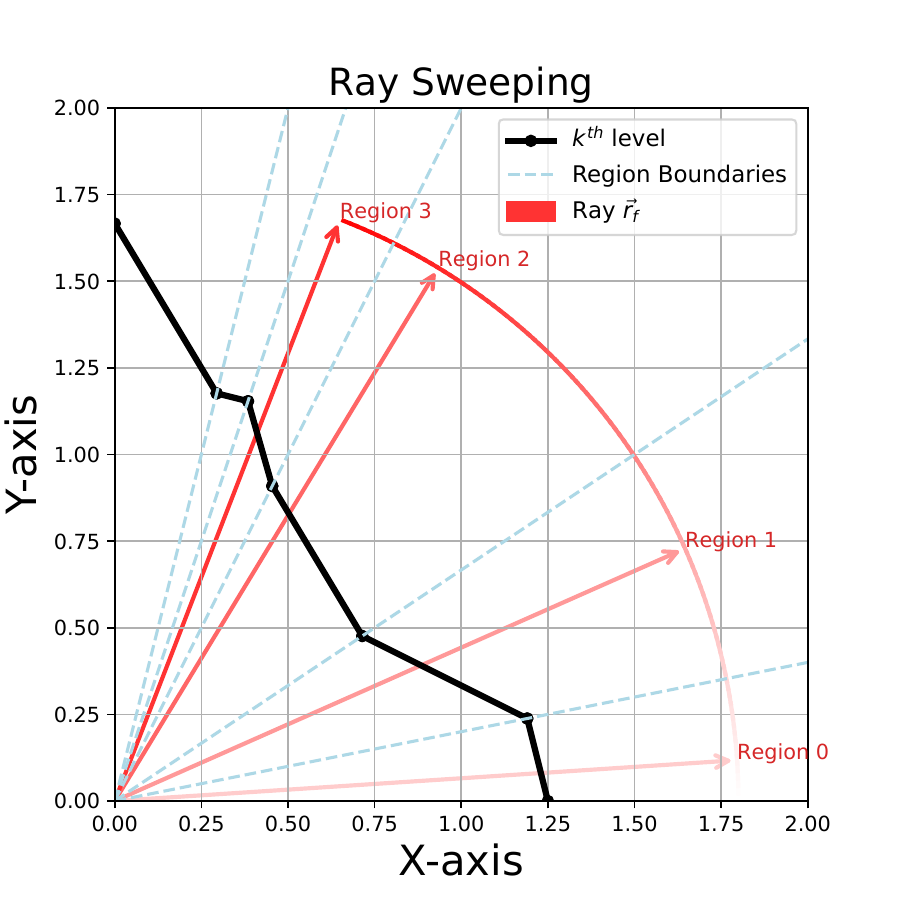}
        \vspace{-5mm}\caption{Sweeping a ray ($\vec{r_f}$) from $x$-axis to $y$-axis counterclockwise.}
        \label{fig:figure2}
    \end{subfigure}
    \vspace{-4mm}
    \caption{A toy example for the Ray-Sweeping algorithm. (a) and (b): Starting from $5$ points in the plane, the algorithm finds the $\frac{n}{2}$-the level of arrangements for the dual lines. (c): Each median region is a line segment in this arrangement. The line segment represents the median point for any direction in this region.
    (d): The algorithm sweeps a ray from the $x$-axis to the $y$-axis for each region, calculates the skew value in this region efficiently, and pushes this value into a heap. In the end, the best directions, in terms of their skew, are extracted from the heap.}
    \label{fig:ray-sweep-toy}
    \vspace{-4mm}
\end{figure*}

The key observation is that while moving between two neighboring median regions, {\em the set of dual lines above/below the median changes by only one element}.
As a result, using the pre-calculated values for the previous vertex, we can calculate the skew value of the current vertex in constant time (except for the very first vertex in the list). Please Refer to  \S~\ref{appendix:aggregation} for more details about the constant-time skew-value update process.
Moreover, when visiting a new vertex in $\mathcal{A}_\frac{n}{2}$, the new median can be found in constant time simply because each vertex is the intersection of the medians of the two neighboring median regions.
The pseudo-code of this approach is provided in Algorithm~\ref{alg:2d:gen_ray_sweep}. 
A visualization of this algorithm on a toy example is provided in Figure~\ref{fig:ray-sweep-toy}.

\vspace{-2mm}
\paragraph{Complexity Analysis}
Steps one and three of the \raysweeping algorithm
are similar to the \warmup Algorithm. For the second part, we calculate each $skew$ value in a constant time. As a result, the running time of the second step is $T_{skew} = O(nk^\frac{1}{3}) = O(n^\frac{4}{3})$. Consequently, the time complexity of the \raysweeping algorithm is no longer dominated by step two. 
Therefore, the time complexity of \raysweeping is equal to the time taken for enumerating the $\frac{n}{2}$-th level of the arrangement -- i.e., $O(n^{\frac{4}{3}} \log^{1 + \alpha} n)$ and $O(n^{\frac{4}{3}})$ for the deterministic and randomized algorithms, respectively.

\vspace{-3mm}
\section{Generalized \raysweeping for Higher Dimensions}

In this section, we extend our \raysweeping algorithm to the general setting, where $d>2$ is a small constant.

\vspace{-2mm}
\subsection{Generalized \raysweeping Algorithm}

For ease of explanation, we explain the \raysweeping algorithm in the first quadrant of $\Reals^d$, where all values are positive.
When $d=3$, the $k$-th level of the arrangement of $n$ hyperplanes has a complexity of the $O(n.k^\frac{3}{2})$~\cite{sharir2000improved}, while its complexity is bounded by $O(n^{\floor*{\frac{d}{2}}} k^{\ceil*{\frac{d}{2}}})$ in general settings~\cite{alon1986number}.
As a result, setting $k=\frac{n}{2}$, $\mathcal{A}_\frac{n}{2}$ has $O(n^{2.5})$ vertices when $d=3$ and at most $O(n^d)$ vertices (each being is the intersection of $d-1$ hyperplanes) when $d>3$.
The SOTA algorithm for enumerating the vertices of the $k$-th level of the arrangement while following the neighboring vertices is a randomized algorithm proposed by Agarwal et al.~\cite{agarwal1994constructing}, with time complexity of $O(n^{\floor*{\frac{d}{2}}} k^{\ceil*{\frac{d}{2}}})$.
Therefore, setting $k=\frac{n}{2}$, enumerating $\mathcal{A}_\frac{n}{2}$ in general setting takes $O(n^d)$.

After enumerating $\mathcal{A}_\frac{n}{2}$, we can follow the same aggregation process as the 2D space and calculate the skew value of each neighboring vertex in time $O(d)$.
As a result, $T_{skew} = O(dn^d)$. Hence, given that $T_{enum} = O(n^d)$, the time complexity of the \raysweeping algorithm in the general setting is bounded by $O(dn^d)$.

The pseudo-code of the \raysweeping algorithm is provided in Algorithm~\ref{alg:2d:gen_ray_sweep}.
The algorithm first constructs the graph of $k$-th level of arrangements, aka the {\em $k$-th level skeleton} of the arrangement. In the skeleton graph, the vertices of $\mathcal{A}_{k}$ are the nodes, while the edges are between the neighboring vertices.
Please refer to Chapter 9 of \cite{edelsbrunner1987algorithms} for more details about the skeleton of an arrangement and its Eulerian tour.
After constructing the skeleton, the \raysweeping algorithm visits its vertices in a depth-first-search manner in $O(n^d)$.

\begin{figure}[t]
\vspace{-3mm}
\begin{algorithm}[H]
\submit{\small}
\caption{Generalized Ray Sweeping}\label{alg:2d:gen_ray_sweep}
\begin{algorithmic}[1]
\Procedure{GeneralizedRaySweeping}{$\dee, \theta$}
\State $\mathcal{H} \gets Dual(\dee)$; $Output \gets [~]$
\State $\mathcal{G}_{\frac{n}{2}} = \mathcal{E}(\mathcal{H})$\Comment{The graph of the k-level skeleton (vertices and edges).}\label{graph_line}
\State $Heap \gets [~]$\Comment{Initialize a max heap.}

\For{$v \in V(\mathcal{G}_{\frac{n}{2}})$ in a DFS traversal order}
\State $f \gets v / ||v||$
\State $skew = GetSkew(\dee, f)$\Comment{Use the aggregated values to compute the skew in constant time.}
\State $Heap.push(\langle f, skew\rangle )$
\EndFor
\While{$Output$ size is less that $l$}
\State $f \gets Heap.pop()$
\State $T \gets \textit{p-}tail(f)$
\If{$L_{T}(\theta) - L_{\dee}(\theta) > \tau$}
$Output.append(f)$
\EndIf
\EndWhile
\State \textbf{return} $Output$
\EndProcedure
\end{algorithmic}
\end{algorithm}
\vspace{-10mm}
\end{figure}

\vspace{-3mm}
\subsection{Constant-time skew-value update process}\label{appendix:aggregation}
In this section, we propose an aggregation method to update the skew of a direction corresponding to a vertex in the $k$-th level of arrangements {\em in constant time} with respect to $n$. Remember the Ray-sweeping algorithm explained in Section ~\ref{sec:2d:ray}. We make a pass over $\mathcal{A}_{\frac{n}{2}}$, which is an ordered list of neighboring vertices in the $k$-th level of arrangement. 

Let $[v_1, v_2, .., v_m]$ be this ordered list of vertices we visit one after the other. 
In other words, 
at each iteration $i\in[m]$, we explore the direction $f_i = \frac{v_i}{|v_i|}$.
Refer to Algorithm~\ref{alg:2d:gen_ray_sweep} for the details. Let \(skew_i = skew(f_i) = \frac{3(\mu_i - \nu_i)}{\sigma_i}\), where
$\mu_i$, $\nu_i$, and $\sigma_i$ are the mean, median, and the standard deviation in the iteration $i$. The mean $\mu_i$ can be calculated in $O(d)$, because $\mu_i = \mu(\dee)^{\top} f_i$, where $\mu(\dee)$ is the mean point of the dataset. We need to calculate and store $\mu(\dee)$ in the pre-processing step. As a result, in the iteration $i$, we can calculate $\mu_i$ in constant time with respect to $n$ (size of $\dee$).

The median $\nu_i$ can be calculated using the vertex $v_i$. Each vertex in the $k$-th level of arrangement is the intersection of multiple adjacent median regions. Each median region corresponds to a segment of a hyperplane, which is the dual transform of the median point for that region. Assume for each vertex in $\mathcal{A}_{\frac{n}{2}}$, we also store the set of median regions intersecting in that point (we use a hashmap to calculate this set of hyper-planes in constant time). This additional information is provided by the algorithms we discussed for constructions of the $k$-th level of arrangement. Now, given a hyperplane $\dual(t)$ in the $k$-th level of arrangement intersecting in $v_i$, we can calculate the median by using the inverse dual transform $\dual^{-1}(\dual(t))$ in $O(d)$ time.

Now let us consider $\sigma_i$:

\vspace{-12mm}
\begin{align}
    \hspace{25mm}\sigma_i = \sqrt{\frac{\sum_{j \leq n} (t_j^{\top} f_i - \mu_i)^2}{n}}
\end{align}

We need to calculate the sum $\sum_j (t_j^{\top} f_i - \mu_i)^2$ in constant time. We can rewrite this as:

\vspace{-4mm}
\begin{align}
    \sum_j (t_j^{\top} f_i - \mu_i)^2 &= \sum_j (t_j^{\top} f_i)^2 + 2 \mu_i \sum_j t_j^{\top} f_i + n \mu_i^2
\end{align}
\vspace{-2mm}

As discussed before, the last term $n \mu_i^2$ can be calculated in constant time. For the second term, we need to store $S = \sum_j t_j$ in the pre-processing step. We can then calculate $2 \mu_i S^{\top} f_i$ in constant time. For the first term, we define $T$ as a $n \times d$ matrix having each $t_j$ as its $j$-th row. Then we can rewrite this as:

\vspace{-4mm}
\begin{align}
    \sum_j (t_j^{\top} f_i)^2 = f_i^{\top} T^{\top} T f_i
\end{align}
\vspace{-2mm}

Where $T^{\prime} = T^{\top} T$ is a $d \times d$ matrix, which can be calculated in the pre-processing step and stored. Then the matrix multiplication $f^{\top} T^{\prime} f$ can be calculated in $O(d)$. As a result, $skew_i$ can be calculated in $O(d)$ -- constant time with respect to $n$. 

\vspace{-3mm}
\section{Practical Heuristics}\label{sec:mdPractical}

The generalized \raysweeping algorithm suffers from the {\em curse of dimensionality}~\cite{edelsbrunner1987algorithms}. 
Therefore, in this section, we propose practical solutions based on search-space discretization (\S~\ref{appendix:discretization}) and smart exploration of the projection space (\S~\ref{appendix:heuristics} and~\ref{high_dim_err_region}).

\vspace{-2mm}
\subsection{MD Discretization}\label{appendix:discretization}
Search-space discretization is common for addressing the curse of dimensionality.
Following this idea, we propose two approaches to discretize the space of all possible directions $f$ in $\Reals^d$ and find the high-skew projections in the discretized space.
Our first strategy is {\em uniform} discretization, where the continuous space of all directions $f$ is partitioned into equal-size grids, and uniformly distributed samples are taken from the grid.
The second strategy works by incrementally building a set of $l$ directions using {\em Quadratic Programming} (QP), where the constraints of the QP are adjusted to enforce exploring the directions farthest from each other.

\vspace{-2mm}
\subsubsection{The Grid-partitioning Approach}
Like the previous sections, we only focus on the positive partition of $\Reals^d$. 
Using the Polar coordinate system, we can uniquely specify each unit vector $f$ in this space by $d - 1$ angels $\{\alpha_1, \alpha_2, \cdots, \alpha_{d-1}\}$ between $f$ and the positive direction of $x_i$ axes ($i \in [d-1]$). Here $\alpha_i \in [0, \frac{\pi}{2}]$. In other words, $f$ comes from an $(d-1)$-dimensional space spanned by $(d-1)$ independent angels.
Now, we can grid-partitioning this space by dividing each angle into equi-size $\epsilon$-length intervals. As a result, the total number of grid-cells would be $(\frac{\pi}{2} \cdot \frac{1}{\epsilon})^{d-1}$. 

According to the problem definition, we want to report top-$l$ high-skew directions $f$ as candidates to find {\yooms}. We should sample at least $l$ directions from the grids to achieve this. We want our samples to span all possible directions, so we should follow a sampling technique to satisfy the diversity of directions.

We follow the DPP (Determinental Point Processing) sampling schema to achieve this ~\cite{anari2022optimal}. Using this approach, one can get a set of $l$ diverse samples from a universe of $n$ points in $O(n \cdot l^{\omega - 1})$, where $\omega \approx 2.37$ is the matrix multiplication factor. 
In this approach, we sample $O(l)$ grids in time of $O((\frac{\pi}{2} \cdot \frac{1}{\epsilon})^{d-1} \cdot l^{\omega - 1})$ and push them in a Max-Heap structure based on the skew value. We analyze each grid by popping the heap and filtering those satisfying the condition of Problem ~\ref{problem}. If we find less than $l$ such directions, we continue sampling more directions (grids). \footnote{The bottleneck of this algorithm is the sampling step.} Calculating the skew of $l$ directions takes an additional $O(l \, n \, d)$ time.

\vspace{-2mm}
\subsubsection{The Candidate-group diversification Approach}
In this approach, we build a diverse set of candidate directions incrementally until we reach the $l$ direction that satisfies the problem requirements. We start from an initial set of unit directions $P_0 = \{u_1, u_2, \cdots, u_d\}$ where $u_j$ is a unit vector lying on the $j$-th axis ($x_j$).

At each iteration $i$, we construct the vector set $P_i$ by adding a new vector $p^{(i)}$ to $P_{i - 1}$. This vector $p^{(i)}$ is as far as possible from the other vectors (alternatively, each vector can also be considered as a point with a unit distance from the origin). We can find such a point by solving the following {\em Quadratic Program}:

\vspace{-6mm}
\begin{align}
    \min& \; y 
    \\ \nonumber 
    s.t. ~&~
    p_j^\top p^{(i)} \leq y \quad \forall j:p_j \in P_{i - 1}
    \\ \nonumber
    &|p^{(i)}| = 1
\end{align}
\vspace{-4mm}

After solving the above QP, we will calculate the skew value for the direction $p^{(i)}$ in $O(d n)$ time. Solving a quadratic program with $n$ variables has a time complexity of $O(n^{3.67} \cdot \log{n} \cdot \frac{1}{\epsilon} \cdot \log{\frac{1}{\epsilon}})$, where $\omega$ is the matrix multiplicative factor $2.37$ and $\epsilon$ is the accuracy. Here the number of variables is $d$, so the running time is $O(d^{3.67} \cdot \log{d} \cdot \frac{1}{\epsilon} \cdot \log{\frac{1}{\epsilon}})$.
As a result, solving the QP $l$ times to get $l$ directions will result in \mbox{$O(l \cdot d^{3.67} \cdot \log{d} \cdot \frac{1}{\epsilon} \cdot \log{\frac{1}{\epsilon}})$}. Calculating the skew on the $l$ directions will take \mbox{$O(l \cdot n \cdot d)$} time. The first part will be the dominant one in terms of time complexity.

\vspace{2mm} 
\submit{
In our technical report~\cite{technicalreport}, we analyze the error bounds of the discretization approaches and provide the following negative result:
let $f^{\prime}$ be a direction with a cosine similarity of (at least) $cos(\alpha)$ with the best direction $f^*$. An upper-bound on skew ratio $\frac{f^{\prime}}{f^*}$ depends on the dataset topology.  
As a result, when discretizing the directions with $\epsilon$-angle spacing, {\em the error bound on the $skew$ value depends on the topological characteristics of the dataset}. 
Still, we guarantee that for each direction $f$ in the top-$\ell$ high-skew local optima (Refer to Problem Definition ~\ref{problem}), we evaluate a direction $f^{\prime}$ such that $cos(f^{\prime}, f) \geq \epsilon$. 
}

\vspace{-3mm}
\subsection{Exploration and Exploitation (E\&E)}\label{appendix:heuristics}

The E\&E heuristic views the problem as a reinforcement learning instance and adapts the exploration-exploitation techniques for finding a set of high-skew directions.  
In particular, our goal is to select the next direction to evaluate sequentially to balance the explore-exploit trade-off.
That is, starting with a small set of randomly selected directions, at each iteration, we either explore the space of possible directions or exploit the best possible directions found so far.

Without loss of generality, we adapt the epsilon-greedy approach~\cite{sutton2018reinforcement}.
Define $\epsilon \leq 1$ as the exploration probability. We flip a coin with probability $\epsilon$ in each iteration and choose to explore or exploit based on this trial result. 
Let $f^{\prime}$ be the best-known solution with the maximum skew so far while considering a hyper-parameter $\tau^{\prime}$ for the cosine similarity of the exploitation phase.
During each exploration step, we draw random directions in the space of all possible directions. On the other hand, 
during the exploitation, we select the best-known direction $f^{\prime}$ and select a random direction from the space of possible direction in the $\tau$-vicinity of $f^\prime$, defined as $\{f^{(s)} | cos(f^{\prime}, f^{(s)}) \leq \tau^{\prime}\}$.
This problem can be reduced to the problem of sampling uniformly from the surface of a unit $d$-dimensional spherical cone with the center $f^\prime$ and the angular radius of $\tau$. We use the sampling approach proposed in \cite{asudeh2018obtaining}.

\vspace{-3mm}
\subsection{Focused Exploration}\label{sec:md:focused}
\label{high_dim_err_region}

A significant challenge for the E\&E approach is that in high dimensional cases where $d\gg 2$, the search space becomes huge, making it unlikely to find near-optimal directions.
Therefore, instead, we first identify the under-performing regions with the hope of reducing the search space significantly.
To do so, we use the model $h_{\theta}$, trained on $\dee$, to find a subset $S\subset\dee$, which we call the {\em error region}, for which $h_{\theta}$ under-performs. Formally, the expected loss of the model in $S$ is significantly higher compared to $\dee \setminus S$.

\begin{figure}[t]
\vspace{-3mm}
\begin{algorithm}[H]
\submit{\small}
\caption{Focused Exploration}\label{alg:md:smart_exploration}
\begin{algorithmic}[1]
\Procedure{SmartExplore}{$\dee, h_{\theta}, \mathcal{K}, \mathcal{K}^{\prime}$}\Comment{$h_{\theta}$ is the trained model on $\dee$.}
\State $\dee_{sorted} \gets sort(\dee, L_{\theta})$\Comment{Sorts $\dee$ based on the loss function $L_{\theta}$ (descending).}
\State $S \gets \textit{First } \mathcal{K} \textit{ portion of } D_{sorted}$\Comment{$S$ is the error region.}
\State $D^S \gets sample(S)$\Comment{Down sample the error region uniformly.}
\State $D^{!S} \gets sample(\dee \setminus S)$
\State $Heap \gets [~]$\Comment{Initialize a max heap.}
\For{all vectors $v \in \{s - t | s \in D^S, t \in D^{!S}\}$}
\State $f \gets v / ||v||$
\State $skew = GetSkew(\dee, f)$
\State $Heap.push(\langle f, skew\rangle)$
\EndFor

\State $Output \gets [~]$
\While{$Output$ size is less that $l$}
\State $f \gets Heap.pop()$
\State $T \gets \textit{p-}tail(f)$
\State $T_{err} \gets \textit{First $\mathcal{K}^{\prime}$ portion of $sort(T, L_{\theta})$}$\label{error_points_of_tail}\Comment{Keep the challenging points of the tail}
\If{$L_{T_{err}}(\theta) - L_{\dee}(\theta) > \tau$}
$Output.append(f)$
\EndIf
\EndWhile
\State \textbf{return} $Output$
\EndProcedure
\end{algorithmic}
\end{algorithm}
\vspace{-10mm}
\end{figure}

After finding the error region, we focus on the directions that separate the error region $S$ from other points $\dee \setminus S$. In other words, we would like to find $f$ such that after projecting $\dee$ on it, the error region is well-separated from others. We start by down-sampling $S$ and $\dee \setminus S$ to explore the specified directions faster.

Given the set of directions pushed into a Max-Heap (similar to \raysweeping), we start popping from the head of the heap and checking the most challenging points in the tail of these directions. Let $\textit{p-}tail(f)$ be the tail of a popped direction $f$ from the heap. For a parameter $\mathcal{K}^{\prime}$, we check the top-$\mathcal{K}^{\prime}$ portion of $\textit{p-}tail(f)$ to evaluation the condition of problem ~\ref{problem}. 

Many implementation details are deferred to \S~\ref{sec:exp} for ease of explanation. A simplified pseudo-code of this approach is given in Algorithm ~\ref{alg:md:smart_exploration}. 
Based on our experiments, {\em this approach works well in practice on large-scale high-dimensional settings.}

\vspace{-3mm}
\section{Experiments}\label{sec:exp}
\vspace{-1mm}

Having presented our algorithms in this section, we evaluate them experimentally for the 2D case and higher dimensions.
We analyze our algorithms in each section and their effectiveness in finding the high-skew directions and candidate \yooms. 
The algorithms are implemented using Python, and the experiments are conducted on a workstation with Ubuntu 18.0 OS.

\vspace{-2mm}
\subsection{2D Experiments}

We ran the 2D experiments on two real-world datasets: (a) {\bf Chicago Crimes} dataset~\cite{chicagocrime2023} and (b) {\bf College admission} dataset~\cite{chand2021admission}.

\vspace{-2mm}
\subsubsection{Chicago Crimes}\label{sec:exp:chicagocrimes}
As stated in \S~\ref{sec:2d}, the 2D algorithms are applicable for datasets with more than two features as long as the projection directions are defined on a pair of attributes of interest.
Therefore, in this experiment, we use the Chicago Crimes dataset~\cite{chicagocrime2023} while using \at{longitude} and \at{latitude} for projection.
This dataset contains the reported crimes in Chicago from 2001 until 2023. It has 22 columns and more than 1.4M rows.

\vspace{-2mm}
\paragraph{Prediction Task}
The target variable in this dataset is binary, reflecting whether a reported crime resulted in an arrest or not.
We trained a simple Neural Network with one hidden layer of size 100 for this dataset.
The model's accuracy on the training dataset is 0.87, with an f1-score of 0.72.

\vspace{-2mm}
\paragraph{Hyper-parameters} In this problem, we first took a sample of size 1000 from the dataset and ran the Ray-sweeping algorithm on the two dimensions, Latitude and Longitude, to find the top high-skew directions with a minimum cosine similarity threshold $cos(\frac{\pi}{12})$. We picked the top-$3$ high-skew directions. After selecting the top directions, we checked the tails of these directions in the original dataset (containing more than 1,400,000 rows). Running the Ray-sweeping algorithm in a sample of size 1000 takes around 15 seconds in the specified system configuration.

To search the entire space of directions, we first normalized the points to lie in the positive direction of the $\Reals^2$ space. In addition to running the Ray-sweeping from the x-axis to the y-axis, we also explored the negative scoped directions. To investigate the negative scoped directions, we transformed the points $(x, y)$ in this space to $(y, A - x)$ for a large value $A$\footnote{This can be viewed as rotating the points 90 degrees counter-clock-wise and renormalizing them.}. This helps us run Ray-sweeping's second pass on the transformed space to cover all negative directions.

\begin{figure}[!tb]
\centering
\begin{subfigure}[t]{.21\textwidth}
  \centering
  \includegraphics[width=\linewidth]{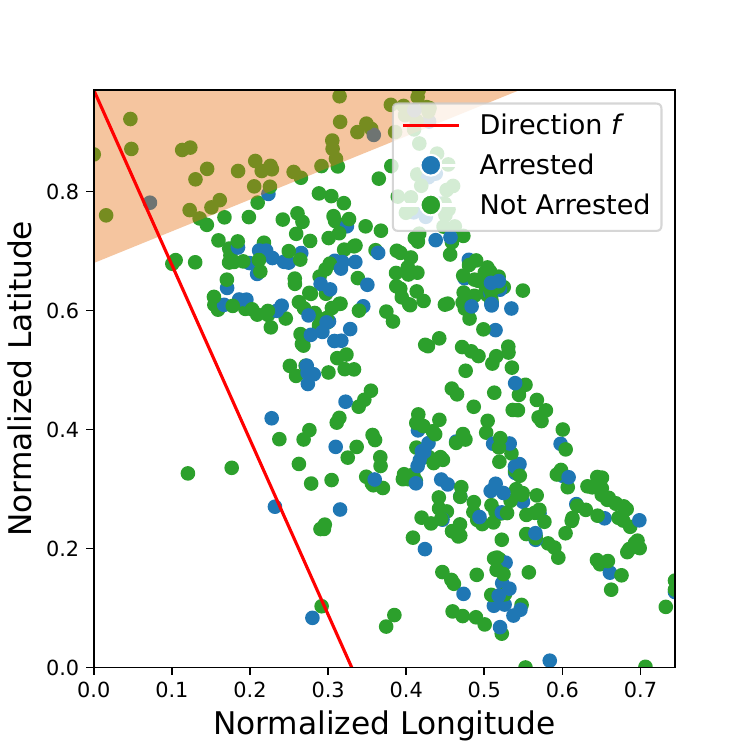} 
  \vspace{-3mm}
  \caption{First high-skew direction.}
  \label{fig:first_skew_2d}
\end{subfigure}
\begin{subfigure}[t]{.23\textwidth}
  \centering
  \vspace{-32mm}
  \includegraphics[width=\linewidth]{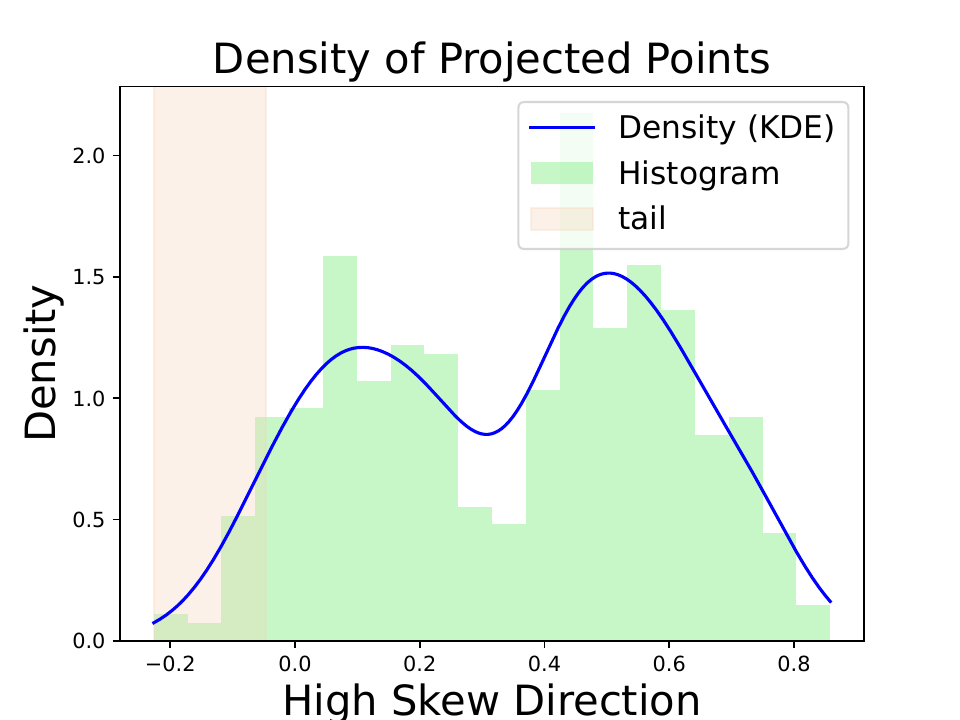} 
  \caption{First high-skew direction. After projecting all the points.}
\end{subfigure}
\vspace{-4mm}
\caption{Top high-skew directions on the Latitude and Longitude of Chicago Crimes dataset are presented in these two plots (the plots show a smaller sample of the dataset points). In the first plot, the output of the binary classification model is presented with green and blue lines. The red line shows the direction $f$ with the highest projection skew. The orange region shows the tail region.
}
\label{fig:skews_2d}
\vspace{-7mm}
\end{figure}

\vspace{-2mm}
\paragraph{Results} 
The top-$2$ high-skew directions are shown in Fig~\ref{fig:skews_2d}. 
As we can see, the first two high-skew directions are along the north side to the south side of Chicago. We select $0.01$, $0.001$, and $0.0001$ percentile of the points in the tail of the high-skew projected points. The tails of the high-skew directions are presented using the orange regions. Evidently, in both cases, {\em the tails correspond to the \underline{North-side Chicago}}. 
To provide some background about the city, Chicago is one of the most segregated cities in the US, with the vast majority of its north side being \at{White}\footnote{Please note that north-side being a \yoom (i.e., under-represented and under-performing) in the Crimes dataset is irrespective of its segregated nature.}. 
We refer the curious reader to \cite{chicagoDemog,redliningChicago} for more details.  
North-side is under-represented in the dataset for many reasons, such as {\em predictive policing}~\cite{daviera2024risk}. 
At the same time, the f1-score significantly decreases on the tail of the projection, illustrating that predictions based on the crime dataset are not accurate for this group. 
To provide more details about these directions, we provide the tail-performance for different percentiles for these two directions in Tables~\ref{tab:first_result} and \ref{tab:second_result}.
As illustrated in the tables, although the accuracy shows only a slight variation, there is a significant decrease in the F1-score on the tails in both cases. This indicates a pronounced class imbalance in the tail distribution, which is attributable to the demographic group imbalance. 

\begin{table}[!tb]
    \centering
    \begin{tabular}{ccl}
    \toprule
    Percentile & Accuracy & F1-score\\
    \midrule
    1        & 0.87     & 0.72      \\
    0.1      & 0.87     & {\bf 0.62}      \\
    0.01     & 0.89     & {\bf 0.68}      \\
    0.001    & {\bf 0.85}     & {\bf 0.40}      \\
    0.0001   & 0.87     & {\bf 0.30}      \\
    \bottomrule
    \end{tabular}
    \caption{(Chicago crime) Performance evaluation on the first high-skew direction (skew = 0.20). The tail is calculated on the entire dataset.}
  \label{tab:first_result}
  \vspace{-8mm}
\end{table}
\begin{table}[!tb]
    \centering
    \vspace{-3mm}
    \begin{tabular}{ccl}
        \toprule
        Percentile & Accuracy & F1-score\\
        \midrule
        1        & 0.87     & 0.72      \\
        0.1      & 0.87     & {\bf 0.62}      \\
        0.01     & 0.89     & {\bf 0.68}      \\
        0.001    & {\bf 0.79}     & {\bf 0.32}      \\
        0.0001   & {\bf 0.81}     & {\bf 0.18}      \\
        \bottomrule
        \end{tabular}
    \caption{(Chicago crime) Performance evaluation on second high-skew direction (skew = 0.16). The tail is calculated on the entire dataset.}
      \label{tab:second_result}
      \vspace{-10mm}
\end{table}



\vspace{-2mm}
\subsubsection{College Admissions}\label{sec:exp:2d:2}
The College-admission dataset ~\cite{chand2021admission} contains details about a group of students who sent applications to US universities. 
This dataset has 400 rows and 7 columns indicating the features of the submitted applications (like GRE, GPA, Gender/Race, etc.). The female-to-male ratio in this dataset is around 1.1. 

\vspace{-2mm}
\paragraph{Prediction Task} The prediction target is whether an application will result in an admission. We used a Logistic Regression model for this binary classification task. It gives us an accuracy of 0.70 and an f1-score of 0.36. We removed the sensitive attribute Gender from the dataset while training this classification model and never used Gender in any implementation steps.

\vspace{-2mm}
\paragraph{Hyper-parameters} We run the 2D Ray-sweeping algorithm on the two columns GRE and GPA. We followed the same process explained for the Chicago crime dataset and investigated the tail of the highest-skew direction.

\vspace{-2mm}
\paragraph{Results} In Fig ~\ref{fig:admission:high_skew}, we provide a visualization of GRE and GPA values from this dataset. The red line shows the highest skew direction. Fig ~\ref{fig:admission:tail} visualizes the tail of this high-skew direction (percentile $p = 0.1$). As we can see in Table ~\ref{tab:second_result_2d}, the ratio of the Female group is significantly larger in the $0.1$, $0.08$, and $0.04$ percentiles compared to the whole dataset (1.1). The accuracy decreases up to \%10 percent when concentrating on the tail of this direction. 
In this case, the Female group is not under-represented in the dataset because the ratio is almost 1.0. However, it is an under-performing and unknown minority group.

\vspace{-2mm}
\paragraph{Runtime} The running time of the Ray-sweeping algorithm on 
the Chicago Crime dataset in our settings was $8$ seconds. This time was $3.74$ seconds for College-admission dataset.

\vspace{-2mm}
\subsubsection{Comparison with Baselines}\label{sec:clustering}
\begin{figure*}[t]
\centering
\begin{subfigure}[t]{.3\textwidth}
  \centering 
    \includegraphics[width=.9\linewidth]{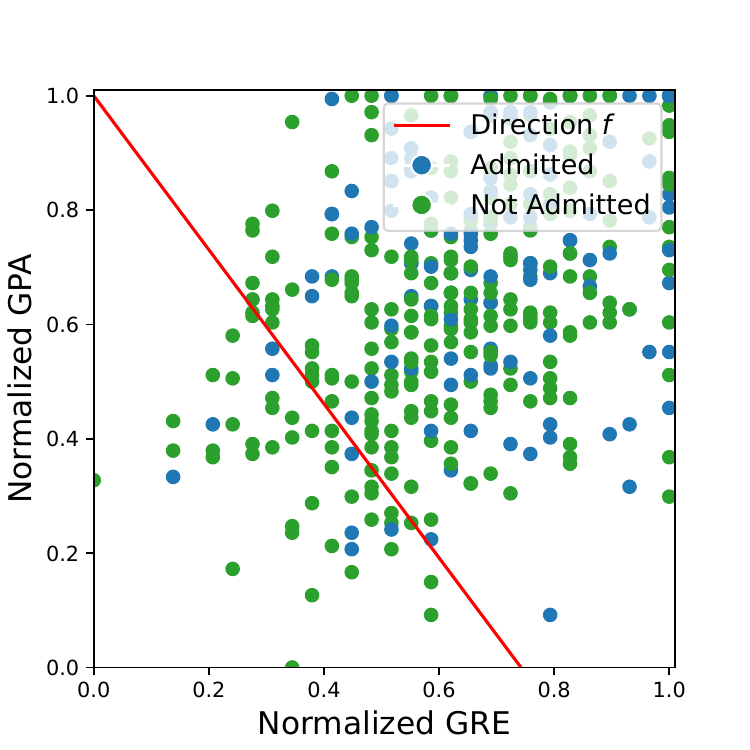}
    \vspace{-3mm}
    \caption{Visualization of GRE and GPA points for the College-admission dataset and the highest skew direction found by the Ray-sweeping algorithm.}
    \label{fig:admission:high_skew}
\end{subfigure}\hfill
\begin{subfigure}[t]{.3\textwidth}
  \centering
    \includegraphics[width=.9\linewidth]{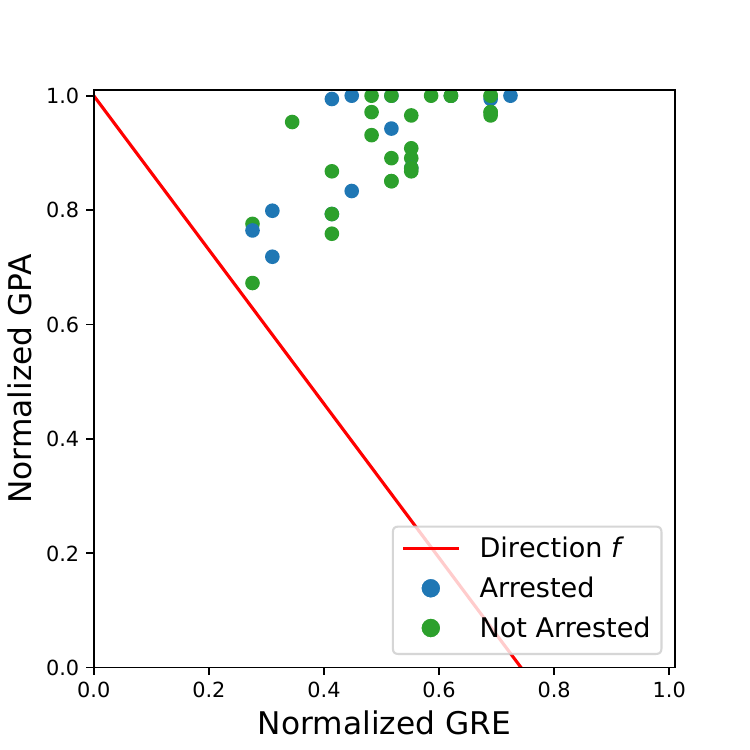}
    \vspace{-3mm}
    \caption{The $0.1-$tail of the highest skew on the College-admission dataset.}
    \label{fig:admission:tail}
\end{subfigure}\hfill
\begin{subfigure}[t]{0.3\textwidth}
\centering
    \includegraphics[width=.9\linewidth]{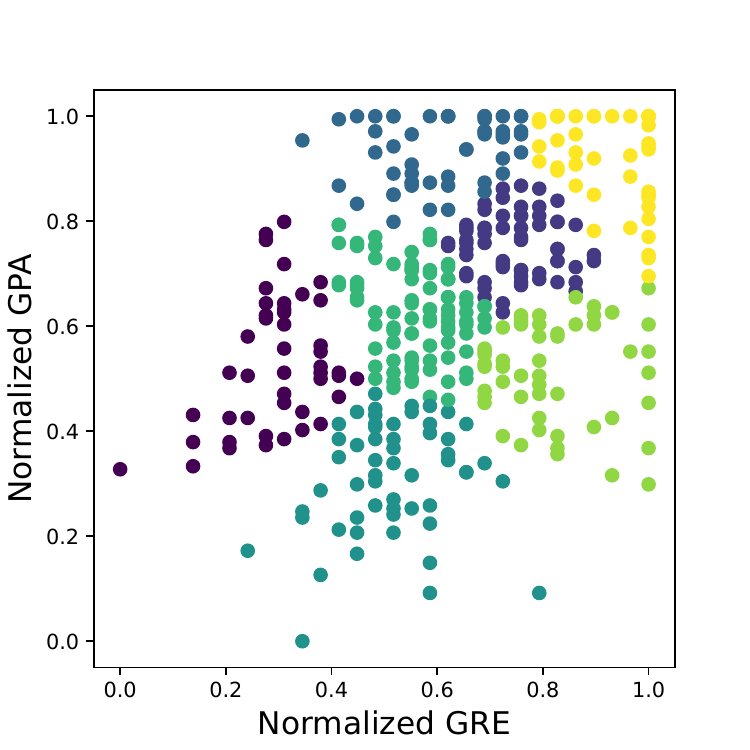}
    \vspace{-3mm}
    \caption{The result of running k-means ($k=5$) clustering on College-admission dataset.}
    \label{fig:admission:kmeans}
\end{subfigure}
    \vspace{-4mm}
\caption{The result of running the Ray-sweeping algorithm on the College-admission dataset.}
\vspace{-3mm}
\end{figure*}

 We utilized two clustering techniques as baselines:
 k-means~\cite{macqueen1967some} and Affinity Propagation~\cite{frey2007clustering}. 
 The resulting clusters were analyzed to identify potential candidates for the underrepresented demographic groups (\yooms). We employed k-means with various values of $k$ and utilized Affinity Propagation with a carefully adjusted \texttt{preference} parameter to control the number of clusters and avoid excessive clustering. A visualization of the k-means clusters on College-admission dataset is provided in Figure~\ref{fig:admission:kmeans}.

The ratio of minority groups within each cluster in k-means for various datasets is presented in Table~\ref{tab:clusters:college}. 
As reflected in the table, unlike the region discovered using our methods (tail of skews), the ratios of minority groups within each cluster are not significantly different from the whole dataset (last row).
As a result, these clusters are not good candidates for identifying {\em underrepresented} groups within the data. Furthermore, the number of points in each cluster is significantly larger compared to the regions we identified, making it more challenging to detect and analyze minority groups within such clusters. In contrast, the regions we discovered are more compact and therefore more efficient for mining {\em underrepresented} groups, offering a practical and focused approach to this task. Table~\ref{tab:ks:college} compares the results for various values of \( k \), all yielding consistent results. 

The results of applying Affinity Propagation clustering are presented in Figure~\ref{fig:cluster:aff:admission}. 
Similar challenges persist, including an excessive number of clusters, requiring all clusters to be examined. Furthermore, the cluster ratios do not significantly differ from those observed in the entire dataset. 

\vspace{-2mm}
\paragraph{Runtime} The average run-time of k-means on the College admission dataset in our setting is $0.003$ milliseconds. 
This value for the Affinity Propagation method is $0.13$ seconds. Even though the running time is smaller for clustering methods compared to the Ray-sweeping algorithm, their output clusters are not valid for solving the Minoria Mining problem defined in this work.

\begin{table*}[!tb]
    \begin{minipage}[t]{0.33\linewidth}
    \centering
    \begin{tabular}{@{}cccc@{}}
        \toprule
        Percen- & Accu- & F1 & Female/Male\\
        tile & racy & score & ratio on Tail\\
    \midrule
    1.00     & 0.70     & 0.36  & 1.10      \\
    0.50     & 0.68     & 0.42  & 1.12      \\
    0.20     & 0.67     & 0.48  & 0.80      \\
    0.10     & {\bf 0.61}     & {\bf 0.34}  & {\bf 2.00}      \\
    0.08     & {\bf 0.64}     & 0.42  & {\bf 1.81}       \\
    \bottomrule
    \end{tabular}
    \caption{Evaluation on the first highest skewed direction (skew = 0.07) on College-admission dataset. The bolded values show a high ratio of minor groups in the tail.}
  \label{tab:second_result_2d}
\end{minipage}
\hfill
\begin{minipage}[t]{0.26\linewidth}
    \centering
    \begin{tabular}{@{}ccc@{}}
        \toprule
    Cluster &  & Female/Male \\
    ID & Size & ratio \\
    \midrule
    0     & 92      & 0.95      \\
    1     & 72      & 0.94      \\
    2     & 108     & 1.11      \\
    3     & 45      & 1.50      \\
    4     & 83      & 1.24      \\
    \midrule
    {\bf Total} & {\bf 400} & {\bf 1.10} \\
    \bottomrule
        \end{tabular}
      \caption{Evaluation of the output clusters of k-means clustering in detecting the unknown minority groups on College-admission dataset with $k = 5$.}
      \label{tab:clusters:college}
\end{minipage}
\hfill
\begin{minipage}[t]{0.36\linewidth}
\vspace{-18mm} 
\centering
    \begin{tabular}{@{}ccc@{}}
        \toprule
    $k$ & Max Size & Maximum F/M ratio \\
    \midrule
    2     & 201     & 1.26      \\
    3     & 136     & 1.20      \\
    4     & 104     & 1.20      \\
    5     & 108     & 1.50      \\
    6     & 90      & 1.44      \\
    7     & 93      & 1.50      \\
    \midrule
    {\bf Total} & {\bf 400} & {\bf 1.10} \\
    \bottomrule
        \end{tabular}
    \caption{Varying $k$ in k-means on College-admission dataset. Col 2 shows the max Female to Male ratio on the output clusters.}
    \label{tab:ks:college}
\end{minipage}
\vspace{-10mm}
\end{table*}

\vspace{-2.5mm}
\subsection{Higher-dimensional Experiments}

In this section, we evaluate the higher-dimensional methods proposed in \S~\ref{sec:mdPractical}. 
As reflected in our negative result on the error-bound of the discretization approaches \techrep{( \S~\ref{sec:md:disc:neg})}\submit{\cite{technicalreport}}, their effectiveness depends on the dataset topology. Besides, their running time significantly increases by the number of dimensions. Our experiments verify these findings: we found discretization slow and less effective. Therefore, in the following we focus on the more practical approaches based on the search space exploration.
Comparing the E\&E method (\S~\ref{appendix:heuristics}) and the focused exploration approach (\S~\ref{sec:md:focused}), our experiments found the latter (focused exploration) significantly more effective for \yoom mining. Therefore, in the following, we first present our results for this approach and then present our extended experiment results for the E\&E approach and highlight some of its drawbacks.

\vspace{-2.5mm}
\subsubsection{Focused Exploration}
We ran experiments on three real-world datasets. The {\bf Adult} Income Dataset ~\cite{Dua2019} consists of demographic information for individuals, from the 1994 Census database. The dataset is primarily used to predict whether a person earns more than \$50,000 per year based on various attributes (14 columns and 48,842 rows). Gender is the sensitive attribute in this dataset. The {\bf COMPAS} dataset ~\cite{angwin2016compas} contains data used to evaluate the effectiveness and fairness of the COMPAS recidivism prediction algorithm. This dataset includes criminal history, jail and prison time, demographics, and COMPAS risk scores for defendants from Broward County, Florida (52 columns and 18,316 rows). Race is the sensitive attribute in this dataset. The last dataset is the {\bf Diabetes} dataset ~\cite{smith1988diabetes}. This is a well-known dataset used for binary classification tasks in the medical domain. It was collected by the National Institute of Diabetes and Digestive and Kidney Diseases and is available through the UCI Machine Learning Repository (47 columns and 101,766 rows). Gender is the sensitive attribute in this dataset.

\vspace{-2.5mm}
\paragraph{Prediction Task} For the Adult income dataset, we trained a Logistic Regression classifier to predict an individual's income (whether it is more than \$50,000 or not) with an accuracy of 0.82 on the training data. For the COMPAS dataset, we trained a simple MLP (Multi-Layer Perceptron) classifier to predict the recidivism score of the individuals (Low, Medium, or High) with an accuracy of 0.71 on the training data. For the Diabetes dataset, we trained an MLP classifier to predict whether a patient is readmitted with an accuracy of 0.70 on the training data. We removed the sensitive attributes from these datasets before training the model, and we never used the information about sensitive attributes in any of our implementations. We only used the sensitive attributes to detect the effectiveness of final high-skew directions in finding \yooms.

\begin{figure*}[t]
\centering
\begin{subfigure}[t]{.32\textwidth}
  \centering
  \includegraphics[width=.9\textwidth]{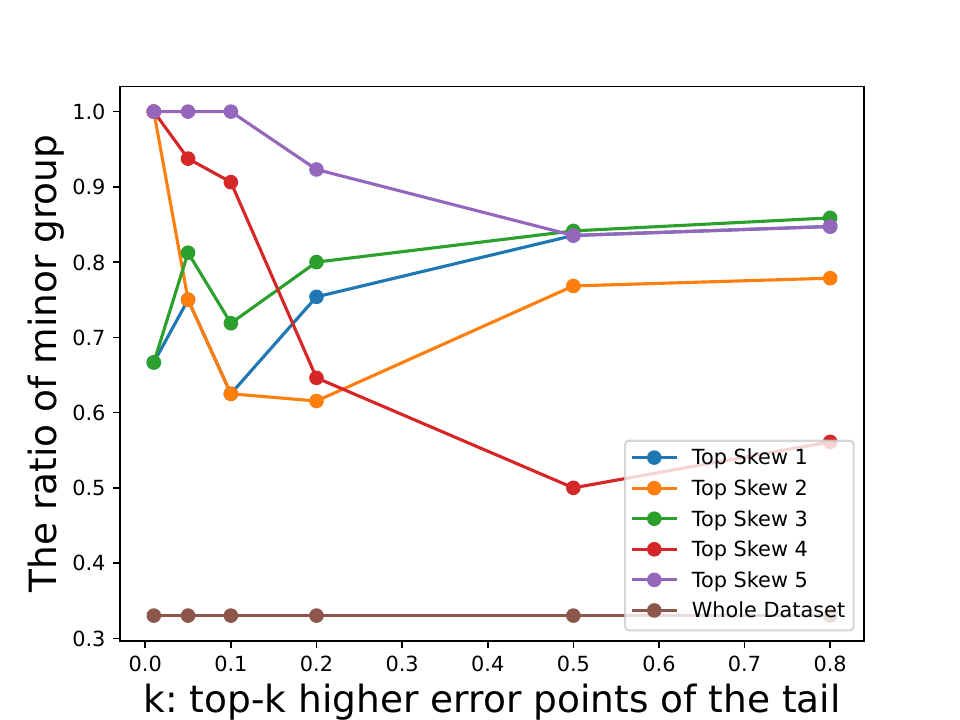} 
  \vspace{-2.5mm}\caption{Adults dataset. Gender is the sensitive attribute.}
  \label{fig:adults_region}
\end{subfigure}\hfill
\begin{subfigure}[t]{.32\textwidth}
  \centering
  \includegraphics[width=.9\textwidth]{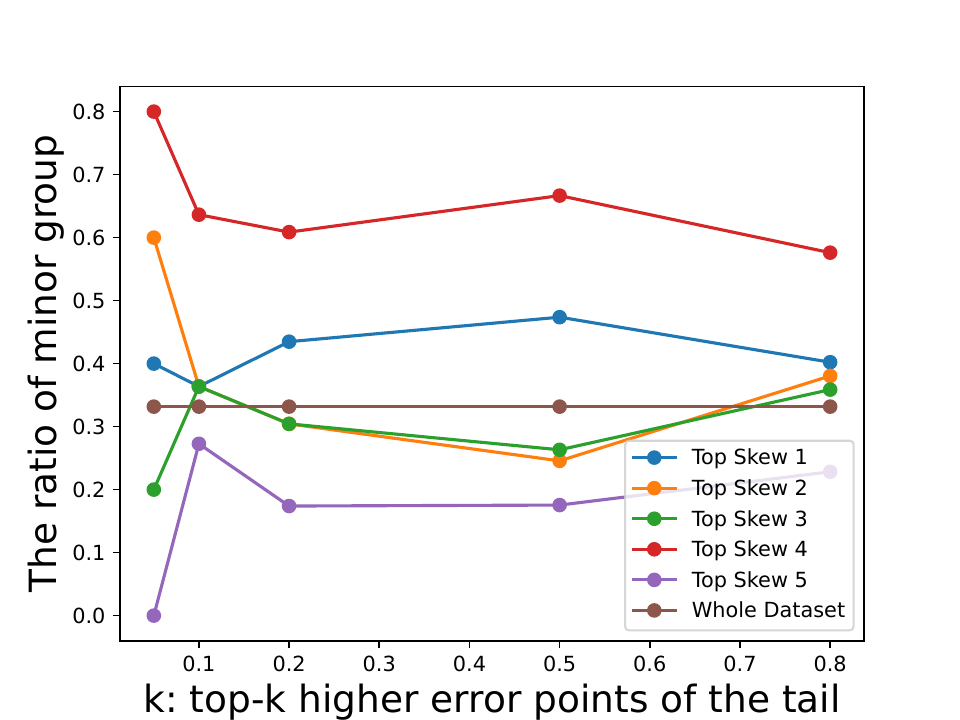}
  \vspace{-2.5mm}\caption{COMPAS dataset. Race is the sensitive attribute (Caucasian vs others)}
  \label{fig:compas_region}
\end{subfigure}\hfill
\begin{subfigure}[t]{.32\textwidth}
  \centering
  \includegraphics[width=.9\textwidth]{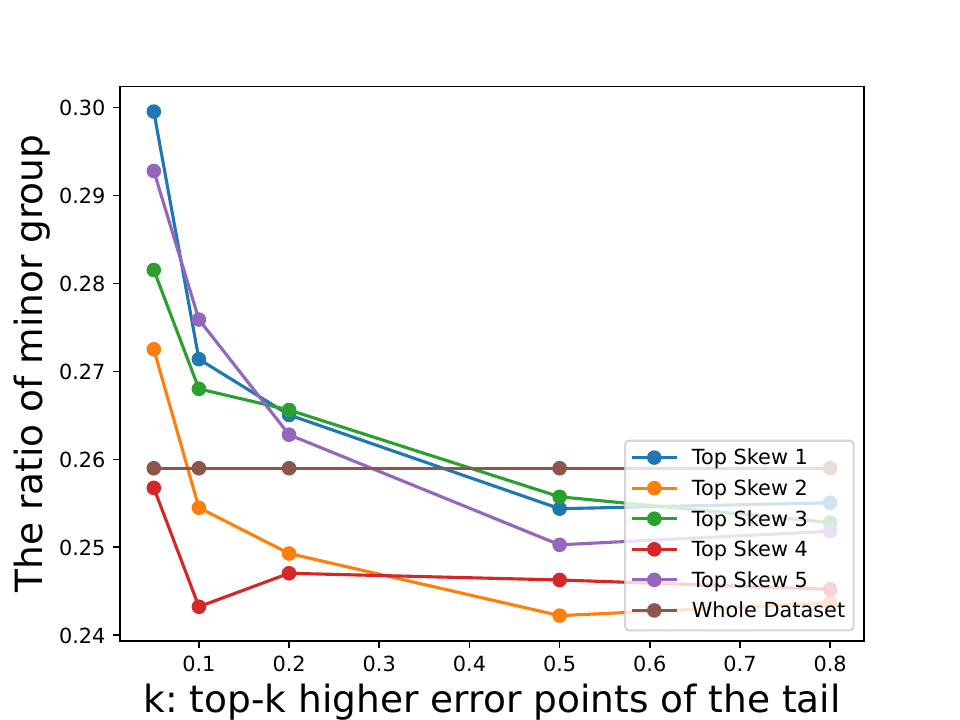}
  \vspace{-2.5mm}\caption{Diabetes dataset. Gender is the sensitive attribute.}
  \label{fig:diabetse_region}
\end{subfigure}
\vspace{-4mm}
\caption{Focused Exploration on the three datasets. Each line corresponds to one of the top-$5$ high-skew directions. The brown line shows the ratio of minor group in the whole dataset. The x-axis shows different values for a hyper-parameter $p$ (or $p$-percentile of the tail). The ratio of minorities (y-axis) is calculated on the top-$k$ (i.e. top $p$-percentile) high error points in the tail of each high-skew direction. If the ratio increases while decreasing $p$, we have more points from the minor group in the set of high error points of the tail of the specific direction.}
\label{fig:focus_explore_result}
\end{figure*}

\vspace{-2.5mm}
\paragraph{Results}
In these experiments, we hide the sensitive attribute containing the under-performing minority group and check if our mining experiments can discover it.
To do so, we solved Problem~\ref{problem} while setting $\ell=5$.
For the Adults and Diabetes datasets, we considered \at{female} as the minor demographic group. For the COMPAS dataset, we considered the race \at{Caucasian} (\at{white}) as the minor demographic group. The result of running the algorithm on the three datasets is shown in Fig ~\ref{fig:focus_explore_result}. For the Adults dataset in Fig ~\ref{fig:adults_region}, three of the five high-skew directions successfully show the Female group as the \yoom. This is because when focusing on the high error points in the tail of these directions, the ratio of Female points increases significantly. In addition, we can see that the ratio of the Female group in the tail of these high-skew directions is larger than this ratio for the whole dataset. This shows that our model did not perform well on the minor group. For the COMPAS dataset in Fig ~\ref{fig:compas_region}, two of the high-skew directions show the Race \at{Caucasian} as the \yoom. The red line is the high-skew direction that shows the model is significantly under-performing on this minor group. For the Diabetes dataset ~\ref{fig:diabetse_region}, all the high-skew directions show that we have more error in the regions where the ratio of the minor group is larger than this ratio in the whole dataset. However, it is not easy for a human expert to detect this minor group as problematic because the ratio of this group is almost less than 0.3 in the high error subset of the tail.
For the Diabetes dataset (Fig ~\ref{fig:diabetse_region}), the tail skewed directions tend to show the problematic points as minor groups. As we focus on the top high error points of the tail of each direction, the ratio of the minor group increases. However, since this ratio is not large enough (less than 0.5), showing the points to a human supervisor may not determine the exact problematic feature in this region.

\vspace{-2mm}
\paragraph{Runtime} The running time of Focused Exploration in our settings 
was $112$ seconds on Adults dataset, $24$ seconds on COMPAS, and $213$ seconds on Diabetes dataset.

\vspace{-2mm}
\subsubsection{Comparison with Baseline}
\begin{figure*}
\begin{subfigure}{0.42\textwidth}
    \centering
    \includegraphics[width=\linewidth]{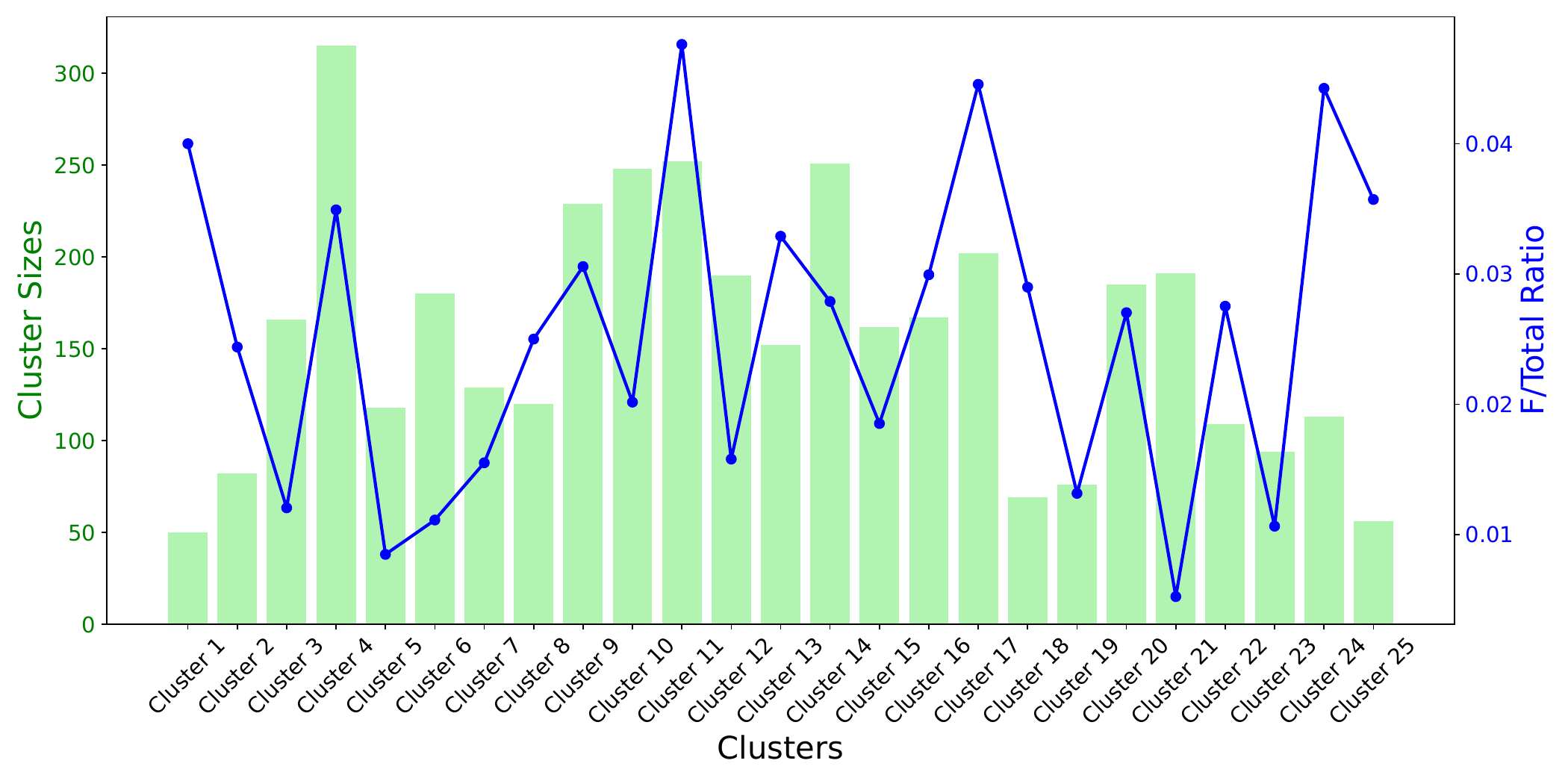}
    \vspace{-7mm}
    \caption{Adults dataset (48,842 points)}
    \label{fig:cluster:aff:adults}
\end{subfigure}
    \hfill
\begin{subfigure}{0.42\textwidth}
\centering
    \includegraphics[width=\linewidth]{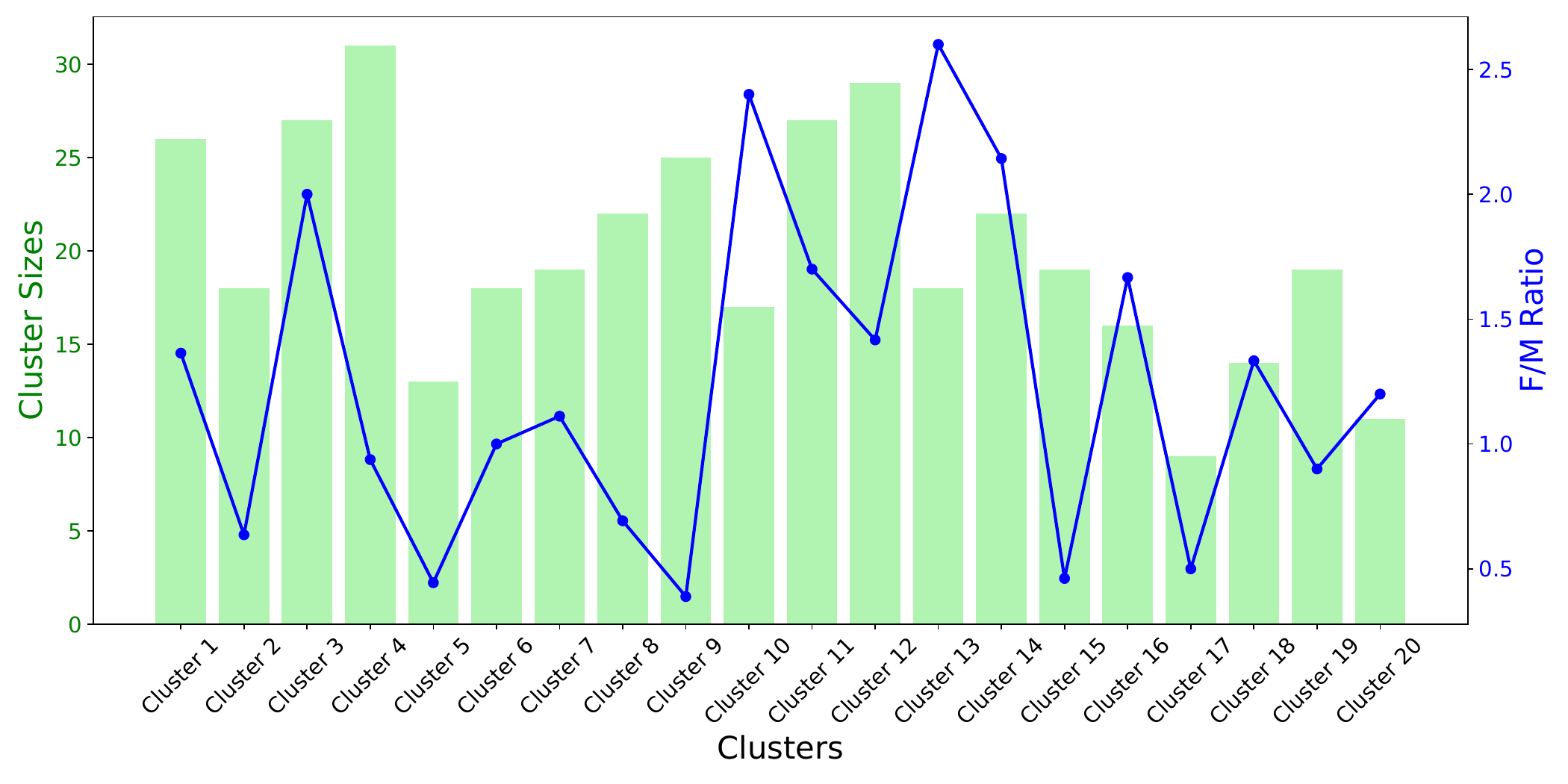}
    \vspace{-7mm}
    \caption{College admission dataset (400 ponints)}
    \label{fig:cluster:aff:admission}
\end{subfigure}
\vspace{-4mm}
\caption{Affinity Propagation clustering: the size and the ratio of the minority group for each cluster are shown in different $y$-axices.}
\vspace{-5.7mm}
\end{figure*}

In this subsection, similar to subsection~\ref{sec:clustering}, we compare the results with two clustering algorithms: 
k-means and Affinity Propagation. Tables~\ref{tab:clusters:adults} and \ref{tab:clusters:compas} show the ratio of minorities on the output clusters of k-means with $k = 5$. Same results as Section~\ref{sec:clustering} is observed here. The result of comparing different values $k$ is shown in Tables~\ref{tab:ks:adults} and \ref{tab:ks:compas}. Applying Affinity Propagation will also result in clusters which are not good candidates for \yooms mining (Figure~\ref{fig:cluster:aff:adults}).

\vspace{-2mm}
\paragraph{Runtime} The running time of k-means clustering on Adults and COMPAS datasets was $0.12$ and $0.33$ seconds, respectively. 
This running time for Affinity Propagation method was $14$ and $20.2$ seconds on Adults and COMPAS datasets. Even though the running time is smaller for clustering methods, their output clusters are not valid for solving the Minoria Mining problem defined in this work.

\begin{table*}[!tb]
    \begin{minipage}[t]{0.22\linewidth}
    \vspace{-3mm} 
    \begin{tabular}{@{}ccc@{}}
        \toprule
        Cluster & Size & female \\
        ID &  & ratio \\
        \midrule
        0     & 10850       & 0.13      \\
        1     & 5718        & 0.30      \\
        2     & 14678       & 0.52      \\
        3     & 3952        & 0.47      \\
        4     & 13644       & 0.24      \\
        \midrule
        {\bf Total} & {\bf 48842} & {\bf 0.33} \\
        \bottomrule
            \end{tabular}
          \caption{Evaluation of the output clusters of k-means when $k = 5$ algorithm in detecting the unknown minority groups for Adults dataset.}
    \label{tab:clusters:adults}
\end{minipage}
\hfill
\begin{minipage}[t]{0.22\linewidth}
\vspace{-3mm} 
    \begin{tabular}{@{}ccc@{}}
        \toprule
    $k$ & Max & Max female \\
    & Size & ratio \\
    \midrule
    2     & 45252       & 0.33      \\
    3     & 19552       & 0.53      \\
    4     & 19522       & 0.52      \\
    5     & 14678       & 0.52      \\
    6     & 13041       & 0.53      \\
    7     & 13670       & 0.84      \\
    \midrule
    {\bf Total} & {\bf 48842} & {\bf 0.33} \\
    \bottomrule
        \end{tabular}
      \caption{Varying $k$ in k-means for Adults dataset. Col 2 shows the max Female/total ratio in output clusters.}
    \label{tab:ks:adults}
\end{minipage}
\hfill
\begin{minipage}[t]{0.22\linewidth}
\vspace{-3mm} 
    \begin{tabular}{@{}ccc@{}}
        \toprule
    Cluster &  Size& Caucas- \\
    ID &  & ian ratio \\
    \midrule
    0     & 524        & 0.09      \\
    1     & 5174       & 0.38      \\
    2     & 5860       & 0.29      \\
    3     & 2781       & 0.44      \\
    4     & 2680       & 0.25      \\
    \midrule
    {\bf Total} & {\bf 18316} & {\bf 0.33} \\
    \bottomrule
        \end{tabular}
      \caption{Evaluation of output clusters of k-means when $k = 5$ algorithm in detecting the unknown minority groups for COMPAS dataset.}
    \label{tab:clusters:compas}
\end{minipage}
\hfill
\begin{minipage}[t]{0.22\linewidth}
\vspace{-3mm} 
    \begin{tabular}{@{}ccc@{}}
        \toprule
    $k$ & Max & Max Cauc- \\
    & Size & asian ratio \\
    \midrule
    2     & 9020       & 0.35      \\
    3     & 7169       & 0.35      \\
    4     & 6638       & 0.45      \\
    5     & 5860       & 0.44      \\
    6     & 5100       & 0.46      \\
    7     & 4511       & 0.46      \\
    \midrule
    {\bf Total} & {\bf 18316} & {\bf 0.33} \\
    \bottomrule
        \end{tabular}
      \caption{Varying $k$ in k-means for COMPAS dataset. Col 2 shows the max Caucasian/total ratio in output clusters.}
    \label{tab:ks:compas}
\end{minipage}
\vspace{-8mm}
\end{table*}

\vspace{-2mm}
\subsubsection{Extended Experiments: the E\&E method} \label{appendix:exp:md}
In this section, we present our experiment results for the E\&E method discussed in \S~\ref{appendix:heuristics} on a synthetic dataset. We aim to evaluate the E\&E method in {\em finding the high-skew directions} of a high-dimensional dataset in different scenarios on the demographic groups. We also mention some observed drawbacks of this method.
We synthesized a dataset to explore the effectiveness of the E\&E approach in finding high-skew directions. This dataset combines two normal distributions on $\Reals^d$ (here, we used $d = 20$) with a total of 20,000 points. Each normal distribution plays the role of a demographic group (for example, male and female groups). The mean and the standard deviation of these two normal distributions are selected in a way that we can explore different scenarios in terms of how well the demographic groups can be separated linearly. After that, we down-sampled one of the regular groups to play the role of an under-represented group in the data. We removed any demographic information from this dataset and ran the E\&E algorithm to find the high-skew directions. 

\vspace{-2.5mm}
\paragraph{Implementation}
We started from 6 different random directions with an exploration probability of 0.4 ($\epsilon = 0.4$); otherwise, we exploited the best-found directions. Refer to  \S~\ref{appendix:heuristics} for more details. We ran 10,000 iterations of E\&E on this synthesized dataset. The running time, on average, was almost equal to 1 minute in the specified system configuration in the previous experiments.

\begin{figure}
    \centering
    \includegraphics[width=.9\linewidth]{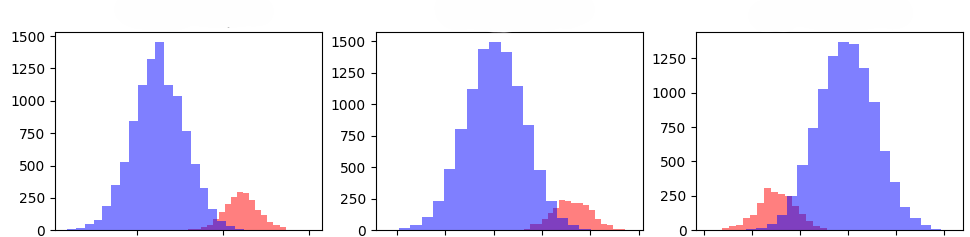}
    \vspace{-3mm}
    \caption{Projected points on highest skewed directions discovered by E\&E. In this case, the demographic groups (hidden during the learning phase) are well separated in the original dataset (the accuracy is almost \%80). The red points belong to the minor group.}
    \label{fig:ee:exp_good_res}
    \vspace{-2mm}
\end{figure}

\begin{figure}
    \centering
    \includegraphics[width=.9\linewidth]{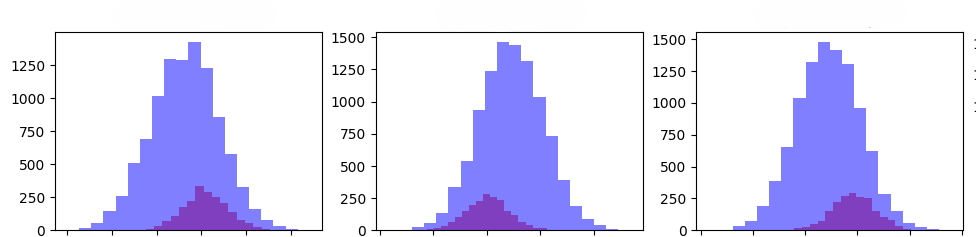}
    \vspace{-3mm}
    \caption{Projected points on highest skewed directions discovered by E\&E. In this case, the demographic groups (hidden during the learning phase) are not linearly separable. The red points belong to the minor group.}
    \label{fig:ee:exp_bad_res}
    \vspace{-5mm}
\end{figure}

\vspace{-2.5mm}
\paragraph{Results}
First, we explored a case when the demographic groups are well-separated linearly. This means there can be a linear classifier to classify these groups with an accuracy of more than \%80. In this case, the E\&E algorithm successfully finds the high-skew directions that separate the minor group from the major one. In Fig ~\ref{fig:ee:exp_good_res}, the top-$3$ high-skew directions are found by this method. The plots show the histogram of the projected points on the high-skew direction $f$. In this case, by picking each direction's tail, we successfully target the minor group (red points). However, consider a case when the two demographic groups are hard to separate linearly (Fig ~\ref{fig:ee:exp_bad_res}). The top-$3$ high-skew directions do not help find the demographic groups in this case. The red points are almost overlapped with the major blue points. As a result, reporting the tail of this skew to a human expert does not help him to detect a \yoom.

The second challenge while using the E\&E approach is the curse of dimensionality. It is hard to find the best highest-skew direction $f^*$ in a high-dimensional case (where $d = 20$). As a result, it will take a long time for this method to converge when the size of a dataset is large (it took 1 minute for a dataset of size 20,000).

\vspace{-3mm}
\section{Related Work}\label{sec:relatedwork}
\vspace{-1mm}


Addressing representation biases in datasets is a well-studied issue in the literature. 
Existing methods study the detection and resolution of representation bias for discrete and continuous data with various modalities~\cite{shahbazi2023representation}.
The literature also includes
statistical and classical methods for identifying patterns and gaps corresponding to empty spaces within the dataset ~\cite{edmonds2003mining,lemley2016big,liu1997discovering}.

Several studies have been conducted to identify the underperforming regions of machine learning models ~\cite{chung2019slice,pastor2021looking,sagadeeva2021sliceline,tae2021slice}. These studies aim to detect regions and sub-spaces that present challenges for the model in making accurate predictions.
SliceFinder~\cite{chung2019slice} detects specific ``slices'' of data where a machine learning model underperforms, indicated by a significantly higher average loss on these slices compared to other parts of the dataset. Other methods build on this concept by concentrating on specific regions of the data ~\cite{sagadeeva2021sliceline}. 
Chen et al.~\cite{chen2018my} employ clustering approaches to identify partitions of data that are discriminated against by a classification model. They aim to discover clusters where the model's performance varies between demographic groups. The issue of identifying underrepresented groups within a sampled population is explored by Lancia et al.~\cite{lancia2024strategies}, who focus on detecting poorly represented groups and resampling these hard-to-sample sub-groups. Pastor et al.~\cite{pastor2021looking} also introduce a concept of pattern divergence to examine a model's behavior across different sub-populations.
Existing work also includes \cite{melika} that combines crowdsourcing and machine learning to detect uncovered groups over predetermined grouping attributes on image datasets without explicit group information.

The primary distinction between our work and the existing literature is that we assume no prior information about the grouping attributes within the dataset. In other words, we aim to mine \yooms from a collection of data points without explicit access to demographic group information over predetermined grouping attributes. Our goal is to identify the groups that are unknown, underrepresented, and underperforming for a given prediction task.
\vspace{-3mm}
\section{Discussion}
\vspace{-1mm}

In this section, we discuss some aspects of our proposed approach and highlight its limitations.

In general, our methods is limited to groups that are (almost) linearly separable. 
If the groups are linearly separable, the high-skewed direction found by our algorithms corresponds to a linear classification.
In cases where linear separation is not feasible,
one can consider introducing new features as polynomial combinations of existing features.

Not all the group candidates identified
as the under-represented and under-performing regions in the data space are socially valid. 
Discovering such regions is intended to help model developers and data owners identify potential issues, decide if those are socially concerning, and {\em proactively}  limit the scope of use of their datasets.


Our approach is only one of the possible solutions for \yoom mining. Finding alternative approaches requires extensive future research. Still, our experiments verified that clustering failed to identify the under-represented and under-represented groups.


Selecting appropriate columns from a dataset to perform Ray-sweeping is a critical practical decision.  
For instance, in our experiments, we utilized the \at{Latitude} and \at{Longitude} columns from the Chicago Crime dataset and the \at{(GRE, GRA)} pair from the College Admission dataset. One potential direction for column selection is identifying the most influential features for black-box machine learning models. Techniques such as feature importance measures or Shapley values can be employed to score each column, allowing the selection of the most impactful features for this algorithm.

In higher dimensions, the curse of dimensionality introduces significant challenges. To address this, we developed heuristics aimed at exploring and exploiting these spaces more effectively. These methods operate at a higher level of abstraction, focusing on identifying high-skew directions. Our findings demonstrate that these approaches, even though just heuristic, outperform traditional clustering methods in finding problematic regions within datasets.


\vspace{-2mm}
\section{Conclusion}\label{sec:conc}
\vspace{-1mm}
In this paper, we introduced, formalized, and studied the problem of mining the unknown, underrepresented, and underperforming minority groups (\yooms).
Making the connection to computational geometry concepts, we proposed efficient algorithms and practical solutions for finding high-skew projections as the combination of data features that reflect high-performance disparities at their tail.
The scope of our proposed techniques is limited to the cases where the \yooms are linearly separable from the rest of the data. Extending our work to the more general class of non-linear projections, as well as developing alternative \yoom mining approaches, is an interesting direction for future work.


\newpage



\balance
\bibliography{ref}

\end{document}